\documentclass{article}

\usepackage[preprint]{corl_2026} 

\usepackage{hyperref}
\usepackage{lipsum}
\usepackage{subcaption}
\usepackage{amsmath,amssymb,amsthm}
\usepackage{graphicx}
\usepackage{multirow}
\usepackage{wrapfig}
\usepackage{mathrsfs}
\usepackage{siunitx}
\usepackage{booktabs}
\usepackage{mathabx}
\usepackage{dsfont}

\newcommand{\tocite}[1]{\textcolor{red}{[CITE]}}

\newcommand{\tocheck}[1]{\textcolor{cyan}{[CHECK]}}

\definecolor{myorange}{RGB}{249,149,51}

\definecolor{DarkBurgundy}{RGB}{128, 0, 32}    
\definecolor{MidnightBlue}{RGB}{0, 51, 102}    
\definecolor{DeepTeal}{RGB}{0, 102, 102}       
\definecolor{SteelBlue}{RGB}{70, 130, 180}     
\definecolor{VibrantPink}{RGB}{225, 40, 130}   
\definecolor{OceanBlue}{RGB}{0, 150, 214}     

\newcommand{\unc}[2]{%
$\begin{array}{c}
#1\\[-3pt]
{\scalebox{0.7}{\tiny$\pm#2$}}
\end{array}$%
}

\newcommand{\uncside}[2]{$#1_{\scalebox{0.6}{$\pm #2$}}$}

\title{Beyond Binary: Sim-to-Real Dexterous Manipulation with Physics-Grounded Contact Representation}

\author{
Jiahe Pan$^{1,2}$ \quad
Stelian Coros$^{1}$ \quad
Jitendra Malik$^{2}$ \quad
Toru Lin$^{2}$ \\
\\
$^{1}$ETH Z\"{u}rich \quad
$^{2}$UC Berkeley 
}

\begin{document}
\maketitle



\vspace{-1em}
\begin{abstract}
A primary bottleneck in contact-rich manipulation is the difficulty of collecting real-world data. Sim-to-real reinforcement learning offers a scalable alternative, but the simulation-reality gap prevents information-dense modalities like touch from being effectively used. Existing sim-to-real methods often mitigate this gap by simplifying tactile data into coarse low-dimensional features -- sacrificing the richness required for complex manipulation. In this work, we introduce \textbf{C}enter-\textbf{o}f-\textbf{P}ressure (\textbf{CoP}), an effective tactile representation grounded in physical principles that preserves dense contact information while maintaining robustness for sim-to-real transfer. To support this representation, we propose a sensor calibration scheme based on differentiable dynamics, enabling the estimation of taxel orientations without requiring ground-truth force measurements. We evaluate CoP on two blind, challenging contact-rich manipulation tasks: peg-in-hole insertion and ball balancing. Across both tasks, policies conditioned on CoP achieve zero-shot sim-to-real transfer on a multi-fingered hand, and outperform both coarse binary-contact and raw-taxel baselines. Analysis of learned policy states further suggests that CoP-conditioned policies encode task-relevant physical properties, such as object mass, as an emergent byproduct of control. \href{https://mpan31415.github.io/tactile_rep/}{\color{VibrantPink}{Project site}} and \href{https://www.youtube.com/watch?v=UXtnP0lUEss}{\color{OceanBlue}{supplementary video}}.
\end{abstract}





\vspace{-0.5em}
\section{Introduction}
\vspace{-0.5em}

A central challenge in robotics is contact-rich dexterous manipulation. While recent advances in imitation learning from large-scale human demonstrations~\cite{bjorck2025gr00t,cheng2025open,zhang2025kinedex} have show impressive progress, the scalability of these approaches is fundamentally limited by the high cost and human effort required for real-world data collection.
Sim-to-real reinforcement learning (RL) offers a compelling alternative through large-scale, autonomous learning in simulation. However, its success has been largely confined to relatively simple tasks~\cite{akkaya2019solving,handa2023dextreme,chen2022system,lin2024twisting,lin2025sim}, where policies can be learned from low-dimensional features that reliably transfer from simulation to the real world. Among common input modalities, tactile inputs are often avoided or heavily simplified, as accurately simulating tactile sensing is notoriously difficult. Tactile sensor responses depend on complex, high-dimensional, and often unmodeled physical processes, and vary widely due to the lack of standardized sensor designs.



As manipulation tasks become more contact-rich and demand greater precision, control policies must reason over more fine-grained physical interactions -- making tactile sensing more essential for policy learning. The lack of an effective tactile representation thus becomes a key bottleneck in extending sim-to-real RL to more challenging settings.
At the core of this bottleneck lies a representation trade-off: simplistic tactile features~\cite{miller2025enhancing,yin2023rotating,qi2023general} enable more robust sim-to-real transfer, while representations closer to raw tactile signals~\cite{higuera2025sparshx,sharma2025sparshskin,yin2025learning,chen2026ptld} contain richer contact information necessary for more complex manipulation.
To balance this trade-off, we introduce \textbf{C}enter-\textbf{o}f-\textbf{P}ressure (\textbf{CoP}), a physics-grounded representation that summarizes local tactile information as a 3D contact force vector and a 3D contact location. CoP provides a middle ground for effective tactile sim-to-real transfer: it is compact enough to align across simulation and hardware, yet expressive enough to preserve force and location information needed for contact-rich dexterous control.
To support this representation, we further propose a sensor calibration method based on differentiable dynamics, enabling the estimation of CoP without requiring ground-truth force measurements.




Existing evaluations can make it difficult to isolate the role of touch, either because policies rely heavily on vision~\cite{chen2023visuo,heng2025vitacformer,sferrazza2024power} or because the tasks involve relatively simple periodic manipulation~\cite{qi2023rma,chen2023visual,yang2025anyrotate,hsieh2025learning}. To address this, we introduce two challenging \textit{blind} manipulation tasks that explicitly require tactile feedback with minimal visual cues: peg-in-hole insertion and ball balancing.
We evaluate our approach through extensive experiments, demonstrating that policies conditioned on CoP achieve successful zero-shot sim-to-real transfer on multi-fingered hands and significantly outperform both simplified and raw tactile baselines. Importantly, we find that learned policies implicitly capture underlying physical properties -- such as object mass -- highlighting the potential of physically grounded representations for enabling more general and scalable manipulation.

\vspace{-0.5em}
\section{Background}
\vspace{-0.5em}

\textbf{Imitation Learning for Dexterous Manipulation}.
Recent advances in teleoperation~\cite{cheng2025open,lin2025learning,wu2024gello,zhao2023learning} and imitation learning~\cite{chi2025diffusion,li2024planning} have enabled remarkable progress in dexterous manipulation.
However, teleoperation data remains costly to scale, and achieving high levels of success and robustness requires large real-world demonstration datasets~\cite{zhao2025aloha,levine2018learning,lin2024data}, making purely supervised methods expensive for reaching human-level dexterity on complex, dynamic tasks.
%

\textbf{Sim-to-Real Transfer of Contact-Informed Policies}.
Sim-to-real RL has demonstrated success for learning manipulation policies in simulation and transferring onto real-world hardware.
Earlier works have used proprioceptive joint control error as an \textit{implicit} representation of external contact~\cite{qi2023rma}; however, the resulting performance gains have been found to be limited~\cite{chen2023sequential}.
The recent emergence of diverse tactile sensors has inspired many \textit{explicit} contact representations, primarily simplified contact representations such as binary~\cite{miller2025enhancing,yin2023rotating} and discretized ternary signals~\cite{qi2023general}.
While these representations can transfer reliably, the coarse discretization discards high-fidelity contact information that is critical for fine-grained contact-rich control.
A few recent works have attempted to directly model and simulate the raw tactile sensor behaviors~\cite{yin2025learning,sferrazza2024power,dang2026hydroshear}, often by training a tactile encoder to compress the raw simulated sensor activations to a latent observation vector for the policy.
Although these approaches better capture the complexity of real-world tactile signals, the resulting representations are often sensor-specific, less interpretable, and harder to align with simulator contact quantities.
A recent work~\cite{yang2025anyrotate} employed a sensor-specific contact representation consisting of force scalar magnitude and contact position in polar coordinates. 
However, a learned tactile encoder is still used, and the evaluation is limited to only in-hand rotation.
%


%
%

\vspace{-0.5em}
\section{Methodology}
\vspace{-0.5em}

\subsection{Physics-Aware Contact Representation}
\begin{figure}[t]
    \centering
    \begin{subfigure}{0.4\textwidth}
        \centering
        \includegraphics[width=\linewidth]{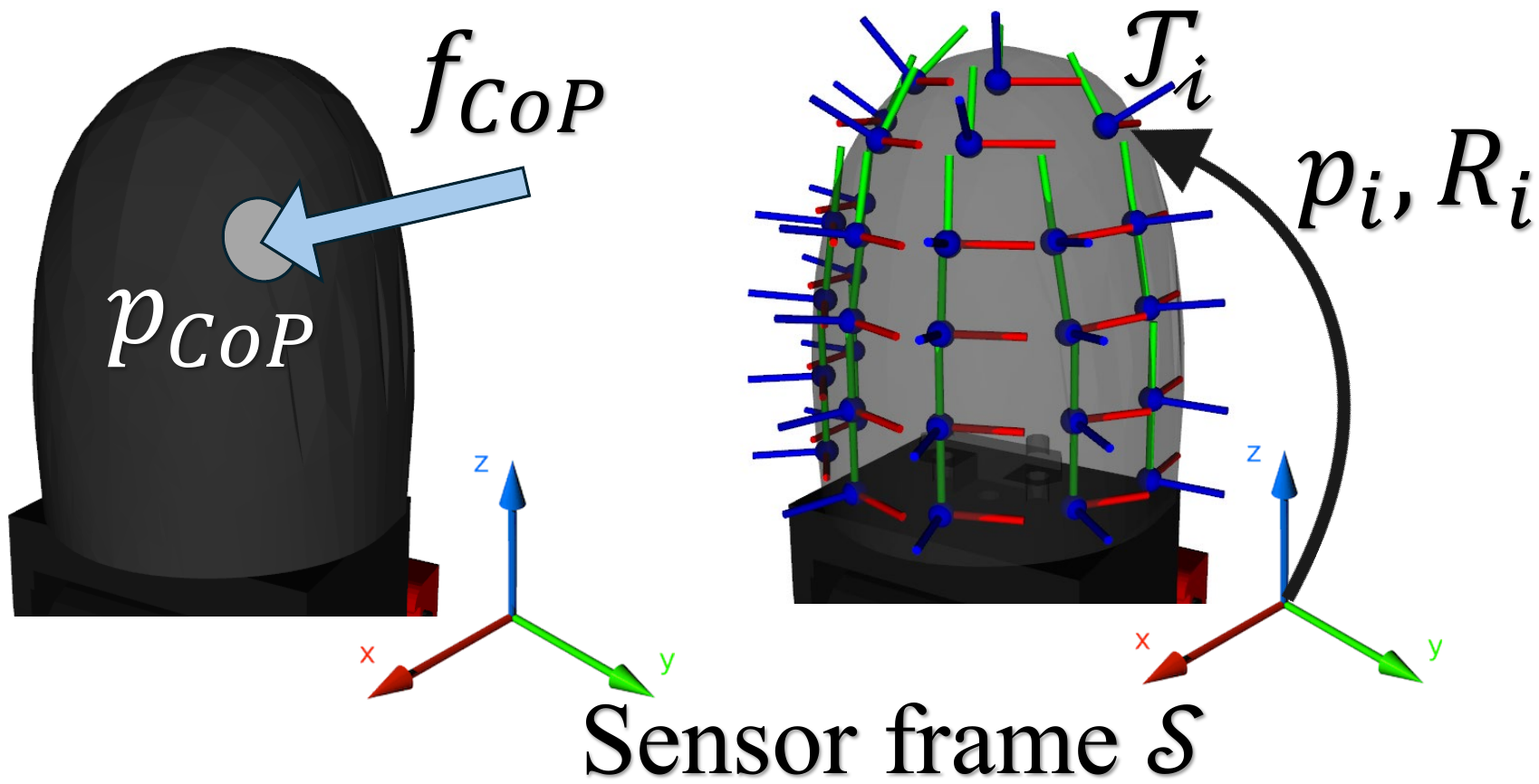}
        \caption{\textbf{Left}: CoP consists of a contact force vector $f_{cop}$ and contact point $p_{cop}$, both expressed in $\mathcal{S}$. \textbf{Right}: Transformation between taxel frame $\mathcal{T}_i$ and sensor frame $\mathcal{S}$.}
        \label{fig:taxel_cop_visualization}
    \end{subfigure}
    \hfill
    \begin{subfigure}{0.59\textwidth}
        \centering
        \includegraphics[width=\linewidth]{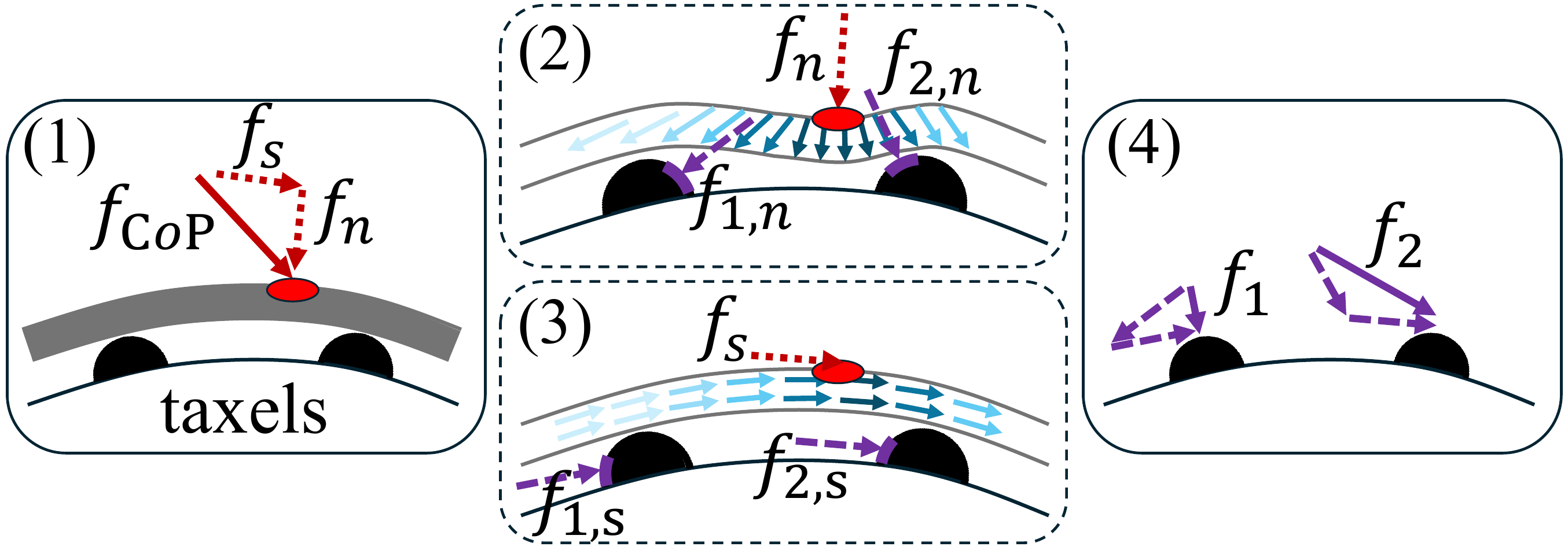}
        \caption{(1) A contact force $f_{\rm{cop}}$ is decomposed into normal $f_n$ and shear $f_s$ components. We model the (2) normal $f_{i,n}$ and (3) shear $f_{i,s}$ forces on each taxel $i$ under surface deformation. (4) Components are combined to obtain the resulting taxel force $f_i$.}
        \label{fig:stress_distribution_model}
    \end{subfigure}
    \caption{(a) CoP representation. (b) The proposed stress distribution model for XELA uSkin sensors~\cite{tomo2016modular}.}
    \label{fig:fingertip_and_deformation}
    \vspace{-0.3cm}
\end{figure}

\vspace{-0.5em}
We propose \textbf{Center-of-Pressure (CoP)}\footnote{We use the term CoP to denote the resultant contact wrench reduced to a single force vector applied at a centroidal contact point; this approximation is intended as a compact local contact descriptor, rather than a complete representation of arbitrary multi-contact pressure distributions.} as a physics-aware contact representation.
On a round fingertip of a robot hand (Fig.~\ref{fig:taxel_cop_visualization}), CoP consists of a 3D force vector $^\mathcal{S}f_{\rm{cop}}\in\mathbb{R}^3$ representing the total contact force acting on the robot link, and a 3D Cartesian contact position $^\mathcal{S}p_{\rm{cop}}\in\mathbb{R}^3$. Both the force vector and the contact point are expressed in the sensor frame $\mathcal{S}$.
Most rigid-body simulators such as IsaacSim~\cite{nvidia_isaac_sim} and MuJoCo~\cite{todorov2012mujoco} provide pairwise contact information between simulation bodies, including 3D contact force vector and the contact position expressed in the world frame. 
%

\subsection{Taxel-CoP Mapping}\label{subsec:taxel_cop_mapping}\vspace{-0.5em}
We show how raw tactile sensor readings can be mapped to CoP representation to enable direct sim-to-real transfer of learned policies, assuming readings are obtained from tactile sensors similar to the XELA uSkin~\cite{tomo2016modular}.
Each sensor array consists of $N$ \textit{taxels} in some grid-like arrangement, and each taxel is an individual tactile sensing point providing 3-axis force measurements.
For the $i$-th taxel, the position and orientation of its local coordinate frame $\mathcal{T}_i$ relative to the sensor frame $\mathcal{S}$ are given by $^\mathcal{S}p_i\in\mathbb{R}^3$ and $R_i\in\mathbb{SO}(3)$ respectively, as shown in Fig.~\ref{fig:taxel_cop_visualization} (right).
Each taxel force $^{\mathcal{T}_i}f_i\in\mathbb{R}^3$ expressed in its local taxel frame $\mathcal{T}_i$ includes both normal and shear components.
We denote the inward, unit surface normal vector at the $i$-th taxel as $^\mathcal{S}\hat{n}_i\in\mathbb{R}^3$, which can be easily obtained as each taxel is arranged such that one of its frame axes is orthogonal to the surface at its position.

Let $^\mathcal{S}f_{\rm{cop}}$ represent an unknown contact force vector on the fingertip. This results in taxel force activations $^{\mathcal{T}_i}f_i$, which are transformed into the sensor frame as $^{\mathcal{S}}f_i$.
We derive a \textit{differentiable bidirectional} mapping between raw taxel forces $\{^{\mathcal{T}_i}f_i\}_{i=1\dots N}$ and the CoP representation $\{^\mathcal{S}f_{\rm{cop}},\,^\mathcal{S}p_{\rm{cop}}\}$ under a parametric stress-distribution model.
From here, since all vector quantities will be defined and expressed in the sensor frame $\mathcal{S}$, we omit frame labeling for better readability.
%


%
\textbf{Stress Distribution Model}.
Directly summing or averaging individual taxel forces~\cite{lee2025trajectory} does not account for force spreading through the compliant silicone layer covering the tactile sensors, and can therefore produce biased estimates of both the resultant force and the contact location.
Instead, we propose a simple model as illustrated in Fig.~\ref{fig:stress_distribution_model}.
First, the CoP force vector $f_{\rm{cop}}$ is decomposed into its normal $f_n$ and shear $f_s$ components.
Then, we model the internal force distribution by accounting for change in force direction due to deformation and force magnitude decay proportional to distance from the contact point.
For each taxel $i$, this results in \textit{effective} normal $f_{i,n}$ and shear $f_{i,s}$ forces, which are combined into the taxel forces $f_i$.
%

%

\textbf{Forward Mapping Derivation}.
We estimate the CoP position $p_{\rm{cop}}$ as the weighted average of individual taxel positions $p_i$ based on their measured force magnitudes $\|f_i\|$. To eliminate idle sensor noise, we include only the active index set $\mathcal{A}=\{i:\|f_i\|>\epsilon\}$ where $\epsilon$ is a threshold value.
To account for the curved fingertip geometry, we approximate the inward unit surface normal at the CoP position, $\hat{n}_{\rm{cop}}\in\mathbb{R}^3$, via inverse distance weighting~\cite{shepard1968two} of the taxel normals $\hat{n}_i$, with weights $\frac{1}{\|p_i - p_{\rm{cop}}\|}$.
%
To model internal stress distribution under deformation, we use a ``blended" unit vector $\hat{b}_i$ to approximate the direction of the effective normal force $f_{i,n}$. Specifically, we interpolate between the taxel's local normal $\hat{n}_i$ and the relative unit direction vector $\hat{v}_i=\frac{p_i-p_{\rm{cop}}}{\|p_i-p_{\rm{cop}}\|}$ from the CoP position to the taxel, $\hat{b}_i = \text{normalize} \left( w_i \hat{n}_i + (1-w_i) \hat{v}_i \right)$, where $w_i$ is a Gaussian radial weight $w_i = \exp(-\frac{\|p_i - p_{\rm{cop}}\|^2}{2\sigma^2})$, and $\sigma$ is a spread hyperparameter.
We approximate the magnitude of $f_{i,n}$ using the same Gaussian decay model on the magnitude of $f_{n}$.
For the effective shear force $f_{i,s}$, we define a shear projection matrix $P_{\text{shear}} = I_3 - \hat{n}_{\rm{cop}}\hat{n}_{\rm{cop}}^\top \in \mathbb{R}^{3\times 3}$ which projects $f_{\rm{cop}}$ onto the surface tangent plane at $p_{\rm{cop}}$ to approximate the direction of $f_{i,s}$, and use the same Gaussian decay model for the magnitude.
Altogether, the relation between $f_i$ and $f_{\rm_{cop}}$ can be compactly written as:
\[
f_i = M_i f_{\rm{cop}}, \quad M_i = w_i (\hat{b}_i \hat{n}_{\rm{cop}}^\top + P_{\text{shear}}) \in\mathbb{R}^{3\times3}
\]
This low-parameter model captures the dominant distance-dependent spreading effect of the stress field while remaining differentiable and easy to align across simulation and hardware.
\textbf{Solving for the CoP Force}.
To find the unknown $f_{\rm{cop}}$ from the observed set of active taxel forces $\{f_i\}$, we aggregate the individual taxel equations into a global linear system $Af_{\rm{cop}} = b$ and obtain $f_{\rm{cop}}$ as the closed-form solution to the following regularized least-squares problem:
\[
f_{\rm{cop}} = (A^\top A + \lambda^2 I)^{-1} A^\top b, \quad A = [M_1^\top, \dots, M_N^\top]^\top, b = [f_1^\top, \dots, f_N^\top]^\top
\]
%
%

With this relatively simple representation, our contact model is not only computationally efficient for policy learning in simulation, but also fully differentiable. This differentiability enables gradient-based learning as presented in Section~\ref{subsec:learning_via_diff_dyn}. Because the same differentiable model can be evaluated in both directions, it enables practical bidirectional conversion between simulated CoP contacts and hardware taxel readings.



\begin{wrapfigure}{r}{0.5\textwidth}
    \centering
    \vspace{-0.4cm}
    \includegraphics[width=0.5\textwidth]{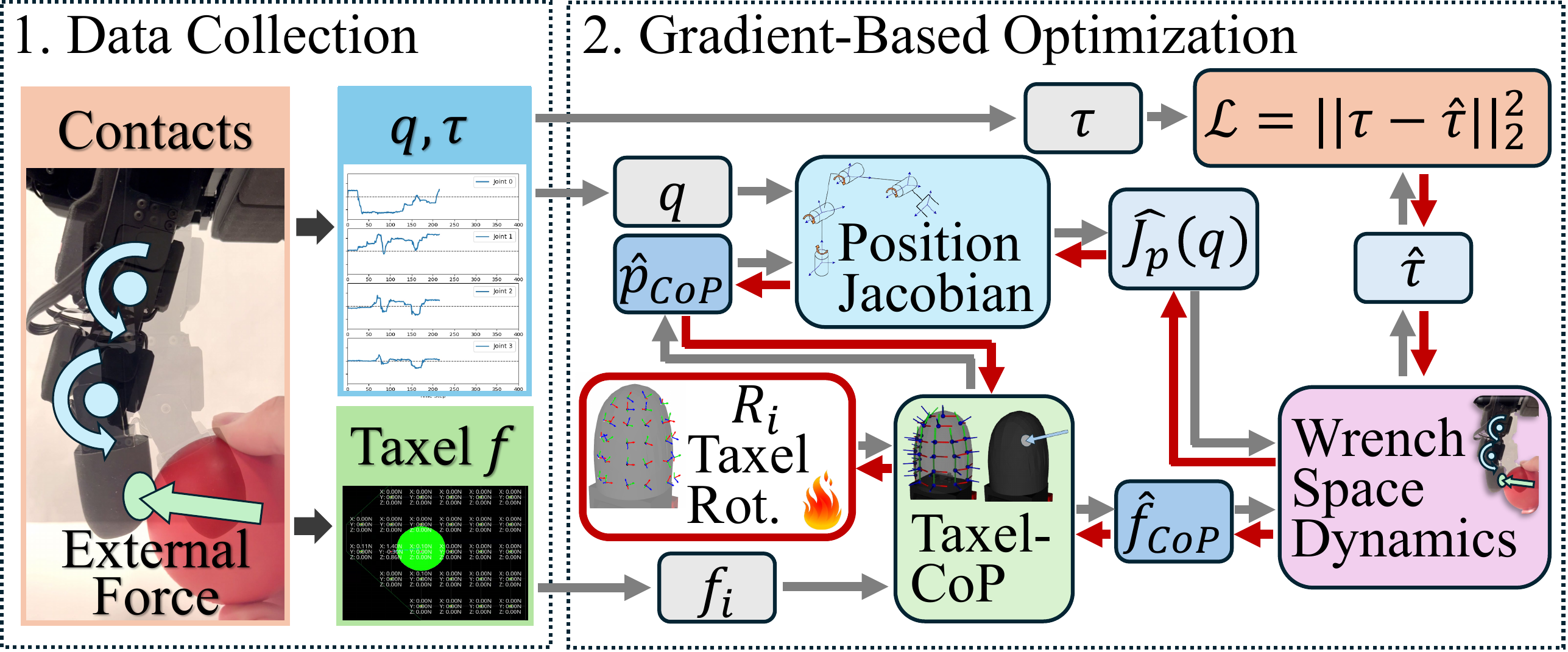}
    \vspace{-0.5cm}
    \caption{Our proposed differentiable dynamics-based sensor calibration method, consisting of 1) data collection and 2) gradient-based optimization. Red arrows indicate gradient flow during back-propagation.}
    \label{fig:learn_via_diff_dyn}
    \vspace{-0.6cm}
\end{wrapfigure}

\subsection{Sensor Calibration via Differentiable Dynamics}\label{subsec:learning_via_diff_dyn}
\vspace{-0.5em}
In the taxel-CoP mapping, while the taxel origins $^\mathcal{S}p_i$ are easily obtainable from the sensor specifications~\cite{tomo2016modular}, the rotations $R_i$ of the taxel frames $\mathcal{T}_i$ relative to the sensor frame $\mathcal{S}$ are unknown and difficult to manually calibrate due to the complex surface geometry of the fingertip.
We therefore estimate the unknown taxel orientations by matching the external wrench inferred from tactile readings to the joint torques required to maintain static equilibrium.
%


%
\textbf{Data Collection}.
Consider the kinematic chain from the robot hand's base frame $\mathcal{B}$ to the sensor frame $\mathcal{S}$ of an arbitrary fingertip.
To collect data, we first run a stiff PD controller continuously to maintain a fixed set of nominal finger joint positions.
Then, we apply random contacts onto the fingertip (see Fig.~\ref{fig:learn_via_diff_dyn}), such that the joint actuators actively apply torques to balance the external force and maintain static equilibrium.
At each time step, we record the raw taxel forces $^{\mathcal{T}_i}f_i$, applied joint torques $\tau\in\mathbb{R}^4$, and joint angles $q\in\mathbb{R}^4$.
To improve identifiability, we apply contacts from diverse fingertip locations and directions, ensuring that the collected dataset excites both normal and shear responses across taxels.

\textbf{Rotation Parameterization}.
For choosing how to parameterize the taxel rotations $R_i$ to be learned, we follow recommendations in~\cite{geist2024learning} and use the $\mathbb{R}^9\mathord{+}\rm{SVD}$ method, which parametrizes a rotation by an arbitrary 3$\times$3 matrix $P\in\mathbb{R}^{3\times 3}$ and projects it to the valid rotation matrix $R \in \mathbb{SO}(3)$ with the least-squares distance to $P$ using its Singular Value Decomposition (SVD):
\[
R = \text{SVD}^+(P) = U\text{diag}(1, 1, \text{det}(UV^\top))V^\top, \quad P = U\Sigma V^\top
\]
We initialize the learnable rotation parameters as a PyTorch tensor of shape $(N, 3, 3)$ consisting of parameter matrices $P_i\in\mathbb{R}^{3\times 3}$ for all $N$ taxels in the sensor array.
\textbf{Optimization}.
During training, for each data sample, we first rotate the recorded taxel forces $^{\mathcal{T}_i}f_i$ into sensor frame $\mathcal{S}$ using the projected rotations $\hat{R}_i$ based on the current $\hat{P}_i$. 
Next, the taxel-to-CoP mapping is applied to obtain the estimated CoP force vector $^\mathcal{S}\hat{f}_{\rm{cop}}$ and contact position $^\mathcal{S}\hat{p}_{\rm{cop}}$, which are transformed into base frame $\mathcal{B}$ via the forward kinematics computed using the recorded joint angles $q$.
Then, we compute the position Jacobian matrix $^\mathcal{B}\hat{J}_{\rm{cop}}\in\mathbb{R}^{3\times 4}$ at the CoP contact point using $q$ and $^\mathcal{B}\hat{p}_{\rm{cop}}$.
Under static equilibrium conditions, the following wrench-space relation with all quantities expressed in a common coordinate frame (e.g. base frame) holds~\cite{huang2011generalized}:
\[
\tau = -J^\top f + g(q) \approx -J^\top f
\]
where $f$ is an external force \textit{acting on} the body, $J$ is the position Jacobian at the contact point, and $g(q)$ is a gravity-compensation term that we assume negligible for simplicity.
Using this relation, we compute the expected joint torques $\hat{\tau}$ required to balance the estimated CoP force vector $^\mathcal{B}\hat{f}_{\rm{cop}}$:
\[
\hat{\tau} = - {}^{\mathcal{B}}\hat{J}_{\mathrm{cop}}^\top
{}^{\mathcal{B}}\hat{f}_{\mathrm{cop}}
\]
Finally, we compute the MSE loss between the estimated $\hat{\tau}$ and recorded joint torques $\tau$, and backpropagate the gradients to obtain improved estimates of rotation parameters $\hat{P}$. Implementation and training details are included in Appendix \ref{appendix:taxel_orientation_learning_implementation}.
%

\subsection{Sim-Real Alignment}
\vspace{-0.5em}

\textbf{Contact Representation}.
We use the $\rm{ContactSensor}$ API in IsaacLab~\cite{mittal2025isaac} to track contacts between the taxel-covered regions of each fingertip link and the contact object.
Because IsaacLab's contact estimates for our fingertip geometries produced unreliable shear components, we use only the surface-normal component of CoP in both simulation and hardware. This choice intentionally sacrifices shear information to maintain sim-to-real robustness.
%
%
To minimize the sim-to-real gap in contact dynamics, we calibrate the taxel-CoP mapping parameters using paired rollout data collected from simulation and hardware. This lightweight calibration is performed once per sensor geometry and does not require task-specific real-world policy learning.
Visualizations of the aligned observations are included in Appendix~\ref{appendix:contact_obs_alignment}.



%

\textbf{Actuator Dynamics}.
Because subtle actuator dynamics (e.g. non-uniform friction) are notoriously difficult to analytically model, we employ a Bayesian optimization-based system identification approach inspired by prior real-to-sim approaches~\cite{lin2025sim,he2025viral} to align the simulated actuator dynamics with the hardware.
Details and visualizations are included in Appendix~\ref{appendix:actuator_sysid}.
%

%

\textbf{Sensor Delay}.
The tactile sensors introduce non-negligible delay from time-of-contact to the policy receiving it as observation, which is critical for highly dynamic tasks.
We measure sensor delay via vision-based methods and introduce it into simulation during policy training.

\vspace{-0.5em}
\section{Experiments}
\vspace{-0.5em}
\label{sec:experiments}

\textbf{Tasks.}
We use a 16-DOF Allegro hand equipped with XELA uSkin sensors covering the fingertips, phalanges, and palm of the hand.
We evaluate on two challenging contact-rich manipulation tasks: \textbf{peg-in-hole insertion} and \textbf{ball-balancing}.
For both tasks, we train a \textit{blind} policy conditioned only on proprioception (current and commanded joint angles) and contact observations.
While existing dexterous manipulation tasks often involve only \textit{primary} contacts (interaction between the hand and the manipulated object), our tasks require the policy to infer \textit{secondary} contacts (interaction between the manipulated object and another task-relevant object).
Success in these tasks crucially depends on the ability of the policy to implicitly infer the state of secondary contacts through the tactile feedback of primary contacts, making them particularly challenging.
%

\textbf{Baselines}.
We compare our proposed CoP representation (\textit{cop}) against several baselines, including proprioception (\textit{base}), binary contact per sensing array (\textit{binary}), and CoP force magnitude (\textit{mag}).
In addition, we include force vector-only (\textit{vec}) and contact position-only (\textit{pos}) as self-ablated baselines, and raw taxel forces (\textit{taxel}).
We also include an expert human (\textit{human}) for comparison.

\textbf{Policy Architecture.}
A common approach of providing temporal information to the policy is to condition it on a flattened stack of past observations $o_{t-N:t}$, which has demonstrated effectiveness in robot manipulation~\cite{miller2025enhancing,sferrazza2024power,qi2023rma,hsieh2025learning}.
We instead use a recurrent policy architecture to provide temporal context without increasing the observation dimension through explicit history stacking.
%
For both tasks, we found that this achieved better sample efficiency and performance over a direct MLP network on stacked history observations (see Appendix~\ref{appendix:policy_arch_comparison} for details).
%

\textbf{Direct Sim-to-Real Transfer.}
We use IsaacLab~\cite{mittal2025isaac} and train policies using asymmetric actor-critic PPO~\cite{schwarke2025rsl}.
The actor observation includes proprioception and the contact representation.
The critic additionally receives privileged information such as object states and other task-relevant information.
Both networks share the same recurrent architecture.
Here, we note that prior works often employ teacher-student distillation~\cite{qi2023general,yin2025learning,chen2026ptld,qi2023rma} because the tactile observations available in simulation and hardware are not directly aligned.
%
In contrast, our aligned CoP representation enables direct sim-to-real transfer.
During training, we systematically introduce domain randomization to robustify the policy against real-world variations. 
Full training details, including observations, actions, rewards, resets, domain randomization, and PPO hyperparameters are included in Appendix~\ref{appendix:rl_training_details}.

\setlength{\tabcolsep}{0.6pt}
\renewcommand{\arraystretch}{1.0}
\begin{table}[t]
\centering
\scriptsize
\caption{Performance comparison of different contact representations for the peg-in-hole insertion task. We report success rate and task completion time (s) across 10 trials for each condition and each insertion shape.}
\label{table:insertion_benchmark}
\begin{tabular}{ccccccccccccc|cc|cccc}
\hline
           & \multicolumn{2}{c}{Circle} & \multicolumn{2}{c}{Diamond}   & \multicolumn{2}{c}{Ellipse} & \multicolumn{2}{c}{Hexagon} & \multicolumn{2}{c}{Square}  & \multicolumn{2}{c|}{Triangle} & \multicolumn{2}{c|}{Overall} & \multicolumn{2}{c}{OOD Init.} & \multicolumn{2}{c}{Masked}   \\
           & sr$\uparrow$         & time$\downarrow$          & sr$\uparrow$           & time$\downarrow$           & sr$\uparrow$           & time$\downarrow$          & sr$\uparrow$          & time$\downarrow$          & sr$\uparrow$         & time$\downarrow$           & sr$\uparrow$           & time$\downarrow$           & sr$\uparrow$            & time$\downarrow$        & sr$\uparrow$            & time$\downarrow$ 
           & sr$\uparrow$            & time$\downarrow$ \\ \hline
           
base       & 0.8        & \unc{5.36}{0.79}          & 0.2          & \unc{4.28}{0.35}            & 0.3          & \unc{2.85}{0.11}           & 0.6         & \unc{2.54}{0.28}           & 0.4        & \unc{2.72}{0.74}           & 0.3          & \unc{10.36}{1.31}           & 0.43          & \unc{4.65}{2.80}       & 0.17            & \unc{11.76}{6.60}          & -              & -     \\ \hline
bin        & 1.0          & \unc{2.60}{0.34}          & 0.2          & \unc{10.82}{3.54}           & 0.3          & \unc{2.58}{0.87}          & 0.8         & \unc{8.30}{3.42}           & 0.6        & \unc{16.73}{4.03}          & 0.3          & \unc{26.29}{0.32}           & 0.53          & \unc{10.15}{8.57}     & 0.20            & \unc{14.40}{5.83}          & 0.52           & \unc{14.87}{17.30}     \\
mag        & 1.0          & \unc{3.10}{0.34}           & 0.5          & \unc{15.94}{7.88}          & 0.4          & \unc{3.95}{1.25}         & 0.8         & \unc{13.03}{7.90}         & 0.5        & \unc{14.08}{10.53}          & 0.1            & \unc{23.50}{0.00}              & 0.55          & \unc{9.47}{9.73}    & 0.27            & \unc{19.33}{7.00}          & 0.48           & \unc{11.13}{16.12}       \\ \hline
vec        & 1.0          & \unc{2.97}{0.64}          & 0.4          & \unc{9.32}{6.54}           & 0.5          & \unc{6.89}{1.30}          & 1.0           & \unc{8.06}{10.29}          & 0.7        & \unc{3.43}{1.98}          & 0.4          & \unc{19.56}{2.60}         & 0.67          & \unc{7.19}{7.60}     & 0.42            & \unc{20.79}{5.07}          & 0.57           & \unc{13.09}{14.58}     \\
pos        & 1.0          & \unc{2.56}{0.30}         & 0.3          & \unc{15.59}{10.96}         & 0.2          & \unc{27.62}{6.31}        & 0.5         & \unc{4.71}{0.31}         & 0.6        & \unc{7.68}{4.28}          & 0.4          & \unc{24.40}{1.00}         & 0.50          & \unc{10.19}{10.12}    & 0.28            & \unc{11.96}{8.01}          & 0.48           & \unc{8.12}{10.53}     \\
taxel      & 0.8        & \unc{9.02}{9.78}         & 0.2            & \unc{20.89}{8.26}              & 0.4          & \unc{5.57}{0.17}         & 0.6         & \unc{2.94}{0.56}         & 0.6        & \unc{17.65}{6.34}         & 0.3          & \unc{18.08}{10.36}         & 0.48          & \unc{10.94}{9.81}     & 0.27            & \unc{15.44}{10.07}          & 0.30           & \unc{9.09}{14.33}    \\ \hline
cop (ours) & \textbf{1.0} & \unc{\textbf{4.22}}{\textbf{0.97}} & \textbf{0.6} & \unc{\textbf{14.89}}{\textbf{7.91}} & \textbf{0.6} & \unc{\textbf{7.61}}{\textbf{4.85}} & \textbf{1.0}  & \unc{\textbf{6.06}}{\textbf{2.99}} & \textbf{0.9} & \unc{\textbf{13.61}}{\textbf{6.11}} & \textbf{0.6} & \unc{\textbf{20.37}}{\textbf{6.39}} & \textbf{0.78} & \unc{\textbf{10.34}}{\textbf{7.62}} & \textbf{0.63}   & \unc{\textbf{15.21}}{\textbf{6.54}}  & \textbf{0.62}  & \unc{\textbf{12.70}}{\textbf{14.71}}  \\
human      & 1.0        & \unc{0.78}{0.19}         & 1.0            & \unc{2.99}{1.13}              & 1.0          & \unc{2.49}{1.32}         & 1.0         & \unc{1.17}{0.30}         & 1.0        & \unc{1.69}{0.94}         & 1.0          & \unc{3.07}{1.19}         & 1.0          & \unc{2.03}{1.32}     & -           & -         & -      & -    \\ \hline

\end{tabular}
\vspace{-0.3cm}
\end{table}
\begin{figure}[t]
    \begin{minipage}[c]{0.34\textwidth}
        \centering
        \begin{subfigure}{\linewidth}
            \centering
            \includegraphics[width=\linewidth]{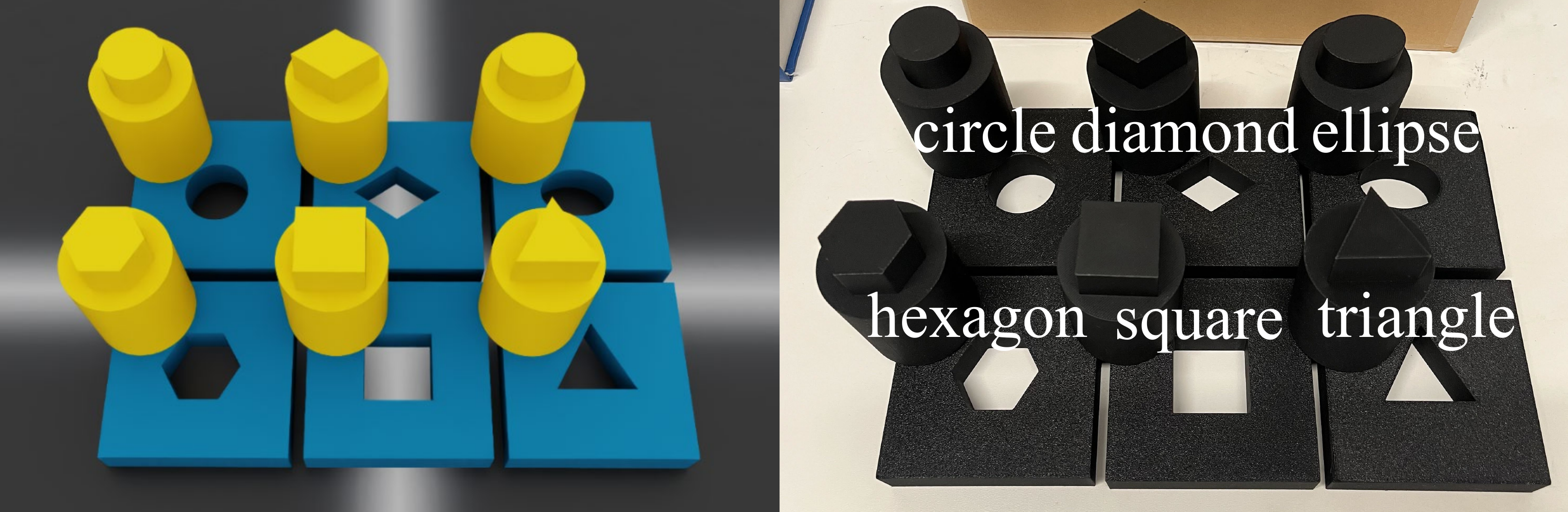}
            \vspace{-0.5cm}
            \caption{Custom insertion task assets.}
            \label{fig:insertion_task_assets}
        \end{subfigure}
        \begin{subfigure}{\linewidth}
            \centering
            \includegraphics[width=\linewidth]{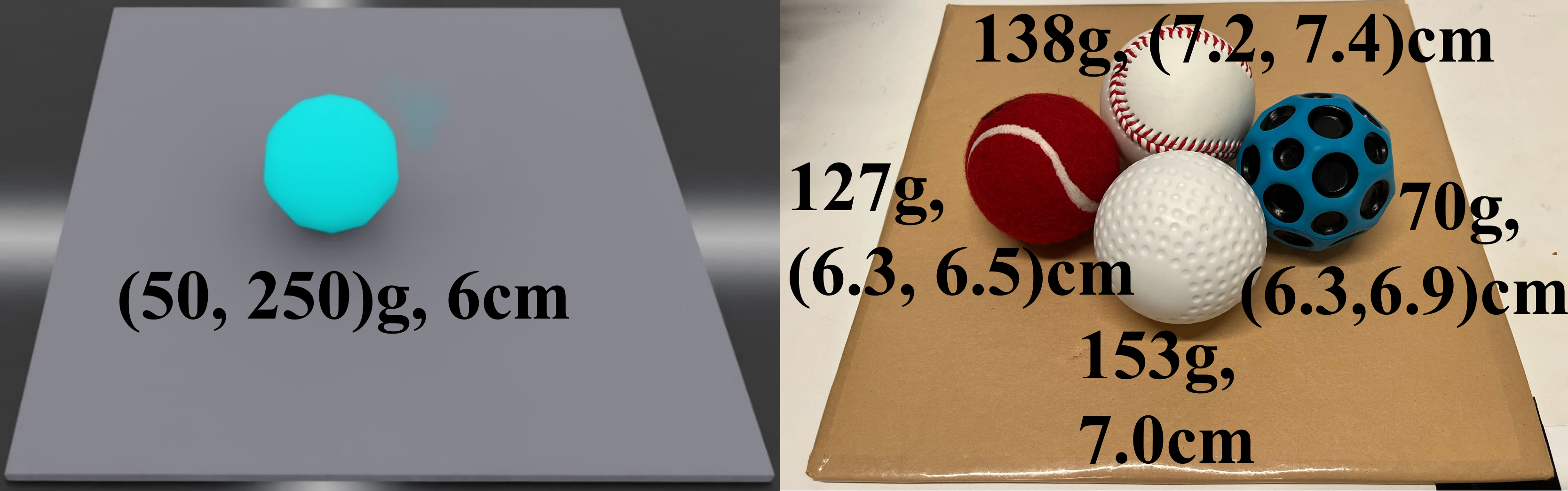}
            \vspace{-0.5cm}
            \caption{Simulation and real-world assets for the ball balancing task.}
            \label{fig:balance_task_assets}
        \end{subfigure}
        \vspace{-0.5cm}
        \caption{Task assets.}
        \label{fig:left_main}
    \end{minipage}
    \hfill
    \begin{minipage}[c]{0.64\textwidth}
        \centering
        \vspace{-0.1cm}
        \includegraphics[width=\linewidth]{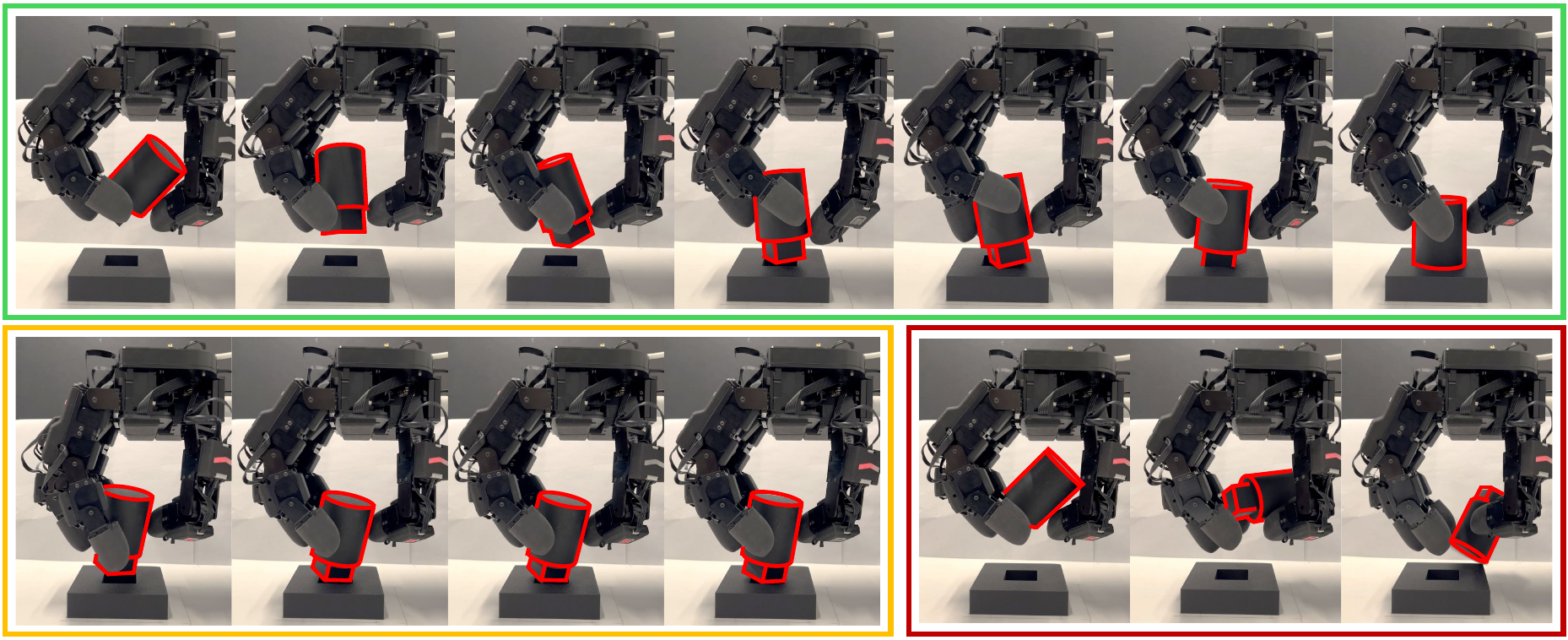}
        \vspace{-0.5cm}
        \caption{\textbf{Top}: The \textit{cop} insertion policy succeeds under OOD initialization. \textbf{Bottom}: The \textit{binary} policy fails under external contacts (left) and OOD initialization (right). The peg object is outlined in red.}
         \label{fig:insertion_real_rollouts}
    \end{minipage}
    \vspace{-0.6cm}
\end{figure}

\vspace{-0.5em}
\subsection{Peg-in-Hole Insertion}
\vspace{-0.5em}
\textbf{Task Setup}.
We generate six pairs of custom peg and hole assets with the same cylindrical handle and a variety of insertion head shapes, as shown in Fig.~\ref{fig:insertion_task_assets} (see Appendix.~\ref{appendix:insertion_asset_details} for asset details).
The task requires the hand to maintain grasp of the peg and fully insert it into the hole object which is fixed to the table.
At each reset, we fully randomize the yaw orientation of the peg, and also introduce small randomizations in its position from manual reset noise.
We conduct 10 trials for each observation and each shape, and record the success rate (sr) and task completion time (time).
\textbf{Real-World Performance}.
As summarized in Table~\ref{table:insertion_benchmark}, \textit{cop} achieves the highest overall success rate and outperforms all baselines on most insertion shapes.
%
Higher-fidelity contact representations (e.g. \textit{vec}, \textit{cop}) generally lead to more adaptive and persistent policies that achieve higher success rates (Fig.~\ref{fig:insertion_real_rollouts}, top) but result in longer task completion times than simplified representations (e.g. \textit{base} and \textit{bin}). On the other hand, simplified representations achieve fast initial insertions only in certain yaw-initializations and fail to adapt to peg-hole contacts and OOD initializations (Fig.~\ref{fig:insertion_real_rollouts}, bottom), suggesting that these representations do not provide sufficient tactile information for recovery once the initial insertion attempt fails.
%
The ablations using force-only (\textit{vec}) and position-only (\textit{pos}) observations indicate that the two components of the CoP representation are complementary and beneficial towards task performance.
Moreover, \textit{taxel} performs worse than nearly all other baselines, possibly due to a combination of imperfect tactile simulation, higher dimensionality, and sensor-specific mismatch.
%
This finding resonates with prior works in tactile simulation for sim-to-real policy transfer, which attempt to alleviate this problem by either training an additional tactile encoder~\cite{yin2025learning} or relying on complex modeling and precise real-world data calibration~\cite{dang2026hydroshear,su2026tacmap}.
Finally, human performance significantly surpasses the best performing robot policy in both metrics.
One possible explanation is that humans combine tactile cues with high-level geometric reasoning and exploratory strategies, whereas the robot policy relies on reactive contact feedback learned from simulation.
%

%
\textbf{Robustness Against Out-of-Distribution Initialization}.
We repeated the experiment under out-of-distribution peg pose initialization (see Fig.~\ref{fig:insertion_real_rollouts}).
As summarized in Table~\ref{table:insertion_benchmark} (OOD Init.), while all policies exhibit longer task completion times as more movements are required to shift the peg downwards towards the hole, \textit{cop} maintains the least reduction in success rate while demonstrating the emergent capabilities of persistent and robust in-hand object translation and re-orientation to achieve peg-hole alignment for successful insertion (Fig.~\ref{fig:insertion_real_rollouts}, top).
\vspace{-0.1cm}

\textbf{Robustness Against Masked Sensor}.
We repeated the experiment while randomly masking 40\% of the raw hardware taxel forces at each time step.  
As shown in Table~\ref{table:insertion_benchmark} (Masked), the higher-fidelity contact representations generally suffer from larger performance degradation than the simplified representations, as the former are more sensitive to precise individual taxel forces.

\begin{figure}[t]
    \centering
    \includegraphics[width=1.0\textwidth]{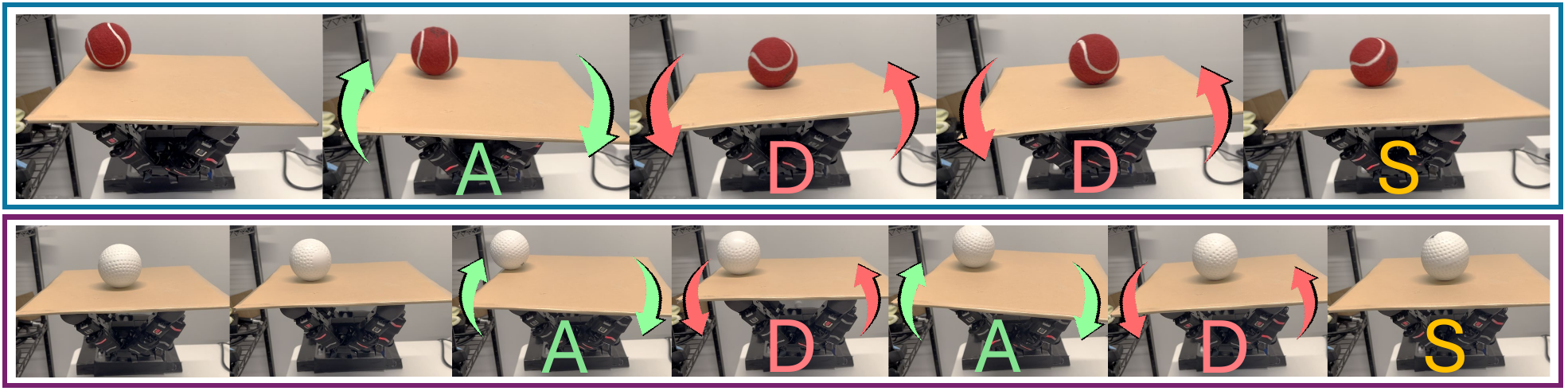}
    \vspace{-0.5cm}
    \caption{Examples of emergent movement patterns of the \textit{cop} policy to center and balance the ball on the plate. ``A", ``D" and ``S" indicate accelerate, decelerate, and stabilize, respectively.}
    \label{fig:balance_real_rollouts}
    \vspace{-0.4cm}
\end{figure}

\vspace{-0.2cm}
\subsection{Ball Balancing}\label{subsec:balancing_task}
\vspace{-0.5em}
\textbf{Task Setup}.
A lightweight (50\si{\gram}) square plate is supported by the four fingertips, and the hand is required to balance a ball placed on the plate and keep it centered as best as possible. 
As shown in Fig.~\ref{fig:balance_task_assets}, the policy is trained in simulation using a smooth sphere and evaluated in the real world on a variety of four balls with different mass, size, friction, and surface texture, resulting in significantly different rolling behaviors outside of the training distribution.
Moreover, this task is particularly challenging for several other reasons.
First, we observed that subtle actuator dynamics such as non-uniform friction and gear backlash are difficult to precisely simulate, which can significantly degrade performance.
Second, one of the fingertips does not have its taxel-covered region in contact with the plate and hence lacks explicit contact information.
Finally, 3/4 of the fingers are close to singularity configurations, making it challenging to achieve rapid fingertip movements.
We conduct 10 trials for each observation and each ball type, and record the time-to-fall (TTF).
\vspace{-0.1cm}

\begin{figure}[t]
  \centering
  \begin{minipage}{0.63\textwidth}
    \centering
    {
\setlength{\tabcolsep}{4.5pt}
\centering
\scriptsize
\vspace{-0.2cm}
\captionof{table}{Performance comparison of different contact representations for the ball balancing task. We report time-to-fall (s) across 10 trials for each condition and each ball type.}
\label{table:balance_benchmark}
\vspace{-0.2cm}
\begin{tabular}{ccccc|c}
\hline
           & Tennis ball$\uparrow$   & Baseball$\uparrow$      & Moon ball$\uparrow$    & \multicolumn{1}{l|}{Hockey ball$\uparrow$} & Overall$\uparrow$      \\ \hline

base       & \uncside{1.38}{0.17}         &   \uncside{1.42}{0.15}        & \uncside{1.50}{0.22}         & \uncside{1.24}{0.16}                            & \uncside{1.38}{0.21}         \\ \hline
bin        & \uncside{2.20}{1.02}          & \uncside{2.22}{1.44}         & \uncside{1.78}{0.71}         & \uncside{1.75}{0.45}                            & \uncside{1.99}{1.03}        \\
mag        &  \uncside{2.83}{0.97}        & \uncside{2.17}{0.48}         & \uncside{2.35}{0.36}        & \uncside{2.25}{0.90}                            &  \uncside{2.40}{0.79}        \\ \hline
vec        &  \uncside{5.59}{3.52}        & \uncside{3.27}{0.41}         & \uncside{4.59}{1.11}        &  \uncside{2.80}{0.91}        & \uncside{4.52}{2.93}        \\
pos        & \uncside{1.63}{0.15}         &  \uncside{1.59}{0.30}         &  \uncside{1.70}{0.16}         &  \uncside{1.26}{0.14}      & \uncside{1.55}{0.27}       \\
taxel      & \uncside{1.38}{0.33}         & \uncside{1.73}{0.31}         & \uncside{1.61}{0.34}        & \uncside{1.22}{0.10}                           &  \uncside{1.49}{0.36}       \\ \hline
cop (ours) &  \uncside{\textbf{5.07}}{\textbf{2.13}} & \uncside{\textbf{4.77}}{\textbf{1.55}} &  \uncside{\textbf{4.50}}{\textbf{1.29}} & \uncside{\textbf{3.06}}{\textbf{0.64}}   & \uncside{\textbf{4.60}}{\textbf{2.19}} \\ 
human      & \uncside{11.29}{6.55}         & \uncside{9.41}{4.46}         & \uncside{5.96}{0.78}        & \uncside{10.82}{5.18}                           &  \uncside{9.37}{5.32}       \\ \hline

\end{tabular}
}
  \end{minipage}
  \hfill
  \begin{minipage}{0.36\textwidth}
    \centering
    \includegraphics[width=\textwidth]{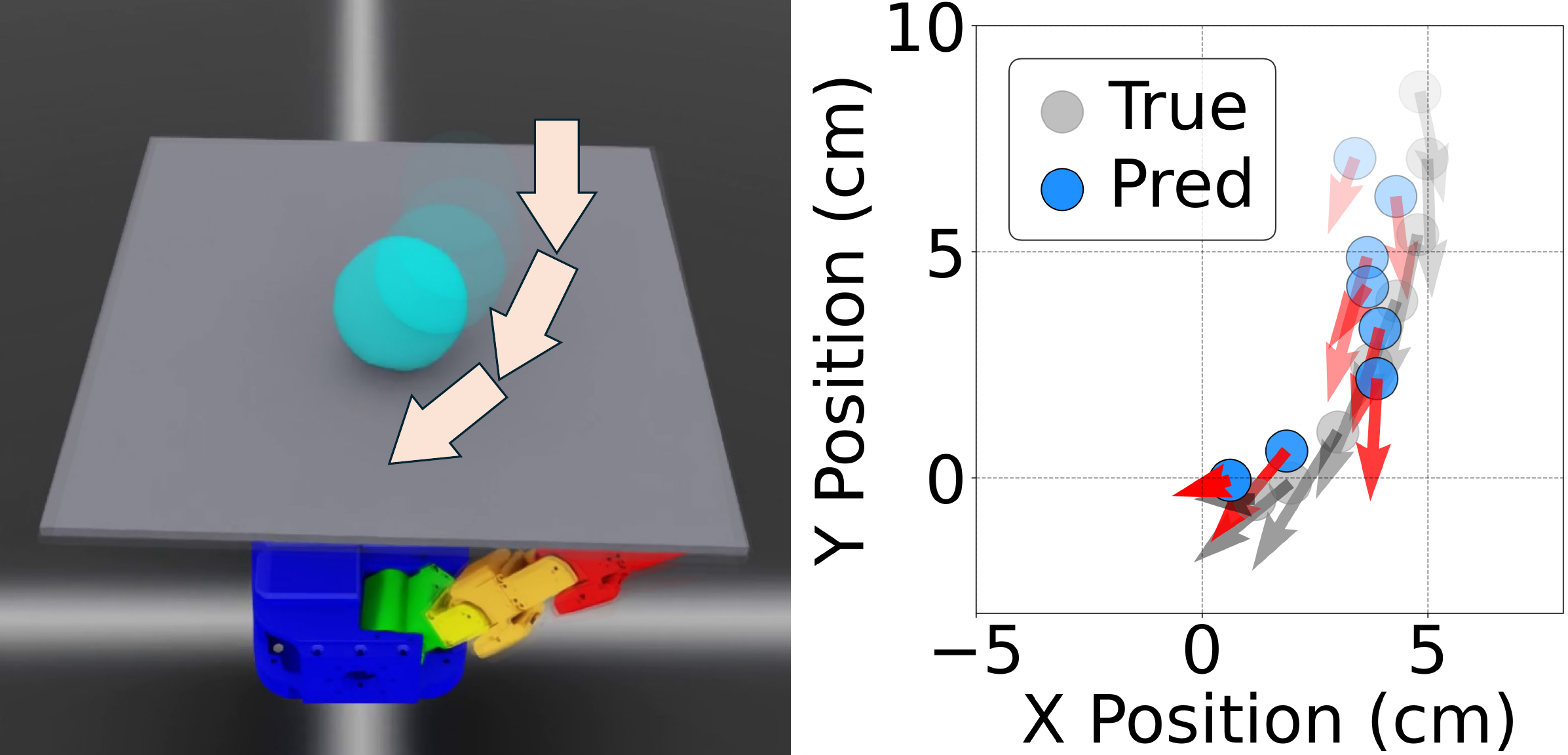}
    \caption{Predicted and actual ball state ($xy$-plane position \& velocity) in a simulated rollout.}
    \label{fig:state_prediction_traj}
  \end{minipage}
  \vspace{-0.7cm}
\end{figure}

\textbf{Real-World Performance}.
As shown in Table~\ref{table:balance_benchmark}, precise force information is crucial for this task, since only \textit{cop}, \textit{vec}, and \textit{taxel} policies, which are conditioned on numerical force values, successfully learned the task in simulation (Fig.~\ref{fig:balance_obs_comparison}).
Moreover, the comparable real-world performance between \textit{cop} and \textit{vec} policies also suggests that force alone may be sufficient for this task.
Fig.~\ref{fig:balance_real_rollouts} shows two distinct emergent movement patterns, including a more aggressive single-step accelerate-decelerate maneuver (top) and a slower two-step centering process (bottom).
Interestingly, while the robot policy struggles more with the smoother and faster-rolling hockey ball, the human expert struggles the most with the dented moon ball which exhibits nonlinear, unpredictable rolling.
As suggested by prior research, one possible explanation is that humans rely more heavily on predictive extrapolation using a \textit{linear} model~\cite{de2017linear,tsutsui2021flexible}.
In contrast, the robot policy may be more reactive towards the immediate state rather than extrapolating future information for control.
\vspace{-0.1cm}

\textbf{Object State Prediction}.
To understand how a trained policy leverages the explicit CoP information for control, we analyze the 256-dim latent output of the recurrent layer of the policy network.
We first collected a training set of 1000 5s trajectories (100 samples each), and performed \textit{linear probing} to predict the ball state consisting of position $p\in\mathbb{R}^2$ and velocity $v\in\mathbb{R}^2$ in the $xy$-plane.
Then, we collected a test set of 100 5s trajectories, and compute the RMSE and $r^2$ scores between the true states and model predictions.
As shown in Table~\ref{table:ball_state_prediction}, the policy's latent representation effectively captures the ball's position.
However, velocity prediction is noticeably weaker, which indicates that while the policy leverages contact information to track ball position, it may not encode motion dynamics as precisely, likely due to the inherent noise in contact-based state estimation.
Fig.~\ref{fig:state_prediction_traj} illustrates the predicted and actual ball states in an example policy rollout trajectory.
\vspace{-0.1cm}

\setlength{\tabcolsep}{2.2pt}
\begin{wraptable}{r}{0.3\textwidth}
\centering
\scriptsize
\captionof{table}{Accuracy of ball state prediction via linear-probing of the policy latents.}
\label{table:ball_state_prediction}
\vspace{-0.2cm}
\begin{tabular}{lllll}
\hline
                           & x pos & y pos & x vel & y vel \\ \hline
RMSE (\si{\meter})                       & 0.013 & 0.019 & 0.059 & 0.065 \\
$r^2$ score & 0.76  & 0.62  & 0.23  & 0.15 \\ \hline
\end{tabular}
\vspace{-0.3cm}
\end{wraptable}

\textbf{Implicit Mass Identification}.
We further examine whether a trained policy's latent representation implicitly captures dynamical properties of the ball such as its mass.
We choose 3 different ball masses (50g, 150g, 250g) and collect 100 5s trajectories for each mass.
Then, we perform PCA to transform all 256-dim latent embeddings into their first two principle components.
%


%
\vspace{-0.2cm}

\begin{figure}[hb]
    \centering
    \includegraphics[width=1.0\textwidth]{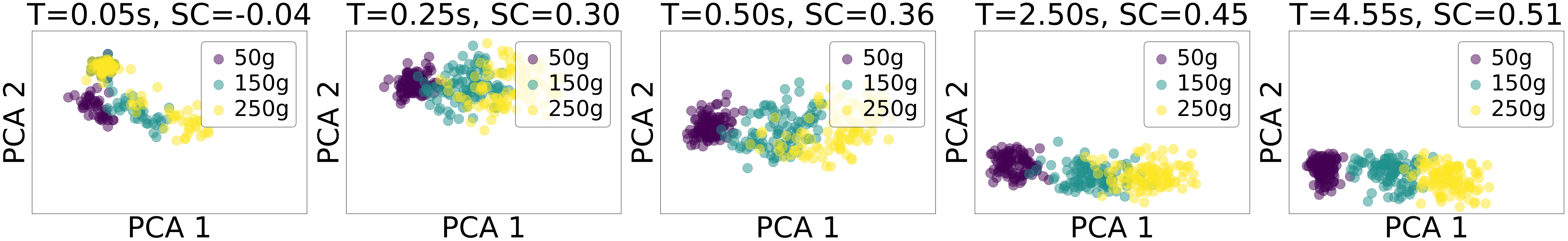}
    \vspace{-0.5cm}
    \caption{Visualization of the emergence of latent embedding clusters across trajectories in temporal evolution, where different clusters correlate with different physical properties. Here we show object mass as an example.}
    \label{fig:mass_clusters}
    \vspace{-0.3cm}
\end{figure}

Fig.~\ref{fig:mass_clusters} reveals that as trajectories evolve over time, the latent embeddings naturally reorganize into distinct clusters corresponding to different mass values, as seen through the gradually increasing Silhouette Coefficient (SC).
This suggests that conditioning on CoP enables the recurrent policy state to organize around task-relevant physical properties, such as object mass, even without explicit supervision.
%
\vspace{-0.1cm}

\vspace{-0.5em}
\section{Conclusion}
\vspace{-0.5em}
We propose Center-of-Pressure (CoP), a physics-aware contact representation that reduces the tactile sim-to-real gap by aligning real taxel readings with contact quantities available in simulation.
We demonstrate its effectiveness through systematic evaluations on dynamic ``blind" manipulation tasks without visual input, and provide insight into the emergent latent representations of the learned policy conditioned on the contact representation.
These results suggest that physically grounded intermediate tactile representations are a promising path toward scalable sim-to-real learning for contact-rich dexterous manipulation.


\vspace{-0.5em}
\section{Limitations}
\vspace{-0.5em}
\textbf{(A) Fidelity vs. Transferability.} CoP intentionally abstracts raw taxel readings into contact force and location information that is more transferable across simulation and hardware. This improves robustness under imperfect tactile simulation, but also discards some sensor-specific details that may be useful for more complex manipulation. Though harder to calibrate and less likely to generalize across sensors, richer raw tactile representations may still offer higher performance when paired with accurate sensor-specific models.


\textbf{(B) Sim-Real Contact Discrepancies.} Our current implementation restricts CoP force vectors to the surface-normal direction because simulated shear force estimates were unreliable. This reflects an important limitation of current contact simulation. In addition, the simulator used in our experiments reports only task-object contacts, whereas real tactile sensors respond to all contacts, including self-collisions and environmental interactions. This mismatch may introduce out-of-distribution tactile observations in more diverse settings.

\textbf{(C) Scope and Future Directions.} We focus on fixed-base dexterous hands with XELA uSkin sensors to isolate the effect of tactile representation. Extending CoP to arm-hand systems, full-hand tactile coverage, and other tactile sensor types remains important future work. Beyond sim-to-real RL, we also hope to integrate CoP as a physically grounded tactile modality for imitation learning and sample-efficient real-world RL.

\clearpage


\bibliography{references}


\clearpage
\appendix
\section*{Appendix}

\section{Taxel Orientation Learning - Implementation \& Training}\label{appendix:taxel_orientation_learning_implementation}
We implement the forward pass calculations based on~\cite{Zhong_PyTorch_Kinematics_2024} which provides differentiable implementations of forward kinematics and Jacobian matrix calculations.
A dataset of 2400 samples (2 minutes at 20\si{\hertz} control rate) containing random object-fingertip contacts was collected.
We used Adam~\cite{kingma2014adam} with a learning rate of $0.1$ and performed batch gradient descent for 100 steps. 
An animation visualizing the training process is provided in the supplementary video.
Prior works mostly use high-precision force sensors to acquire ground-truth data for sensor calibration~\cite{yang2025anyrotate,dang2026hydroshear,freud2025simshear,choi2025coinft,chelly2025tactile,kim2025robust}.
Instead, our method leverages the differentiability of robot dynamics and our proposed taxel-CoP mapping.
Moreover, the framework can in theory learn any differentiable mapping function (e.g. neural network) from raw sensor readings to the CoP representation.
%

\section{Contact Observation Alignment}\label{appendix:contact_obs_alignment}
\begin{figure}[h]
    \centering
    \includegraphics[width=1.0\textwidth]{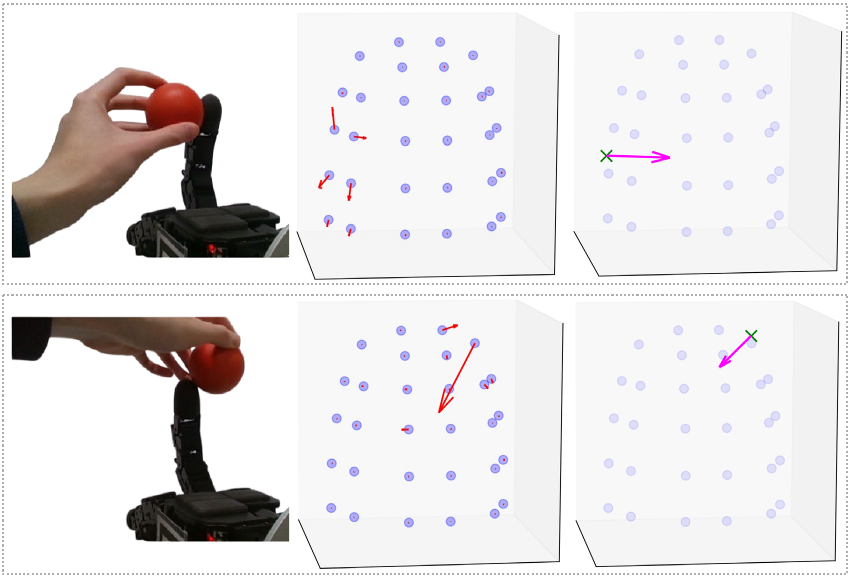}
    \caption{Two example visualizations of the computed CoP contact representations (green cross \& pink arrow) from raw taxel forces (red arrows) on hardware. In each example, the middle subplot shows the raw taxel force vectors, and the right subplot shows the computed CoP force vector and contact position.}
    \label{fig:real_taxel_to_cop}
\end{figure}

\begin{figure}[h]
    \centering
    \includegraphics[width=1.0\textwidth]{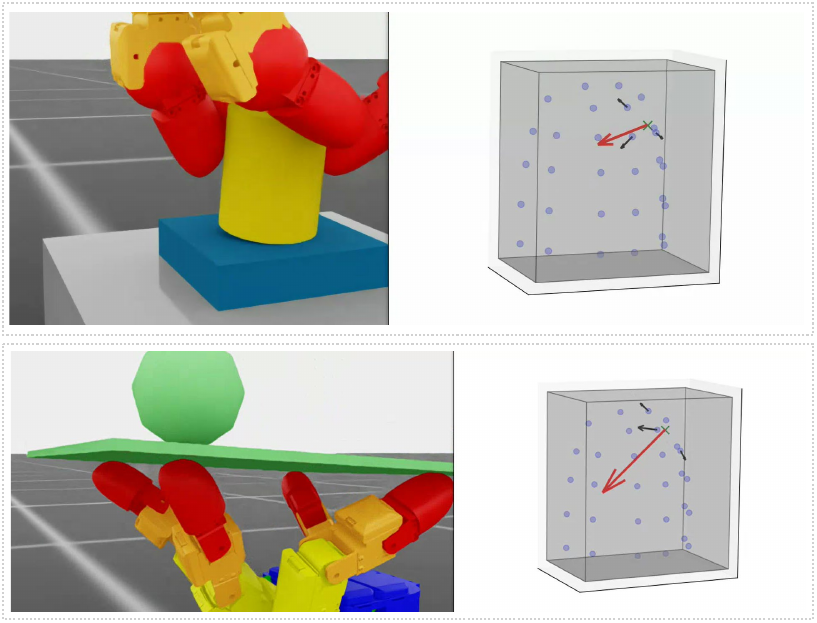}
    \caption{Visualizations of the computed taxel force vectors (black arrows) from contacts on the \textit{middle fingertip} given in CoP representation (green cross \& red arrow) from the simulator, for the insertion (left) and ball-balancing (right) tasks. Grey boxes represent the taxel-covered regions which are used to filter contacts during simulation.}
    \label{fig:sim_cop_to_taxel}
\end{figure}

The proposed bidirectional taxel-CoP mapping enables us to both convert raw taxel activations into the CoP contact representation and compute taxel forces from a CoP contact interaction.
Fig.~\ref{fig:real_taxel_to_cop} visualizes the computed CoP force vector and contact position from raw taxel forces recorded on hardware during real-world contact interactions with an external object.
Fig.~\ref{fig:sim_cop_to_taxel} visualizes the CoP contact information provided by the simulator and the computed raw taxel force vectors using the taxel-CoP mapping, where both show the forces on the middle fingertip as an example.


\begin{figure}[h]
    \centering
    \includegraphics[width=1.0\textwidth]{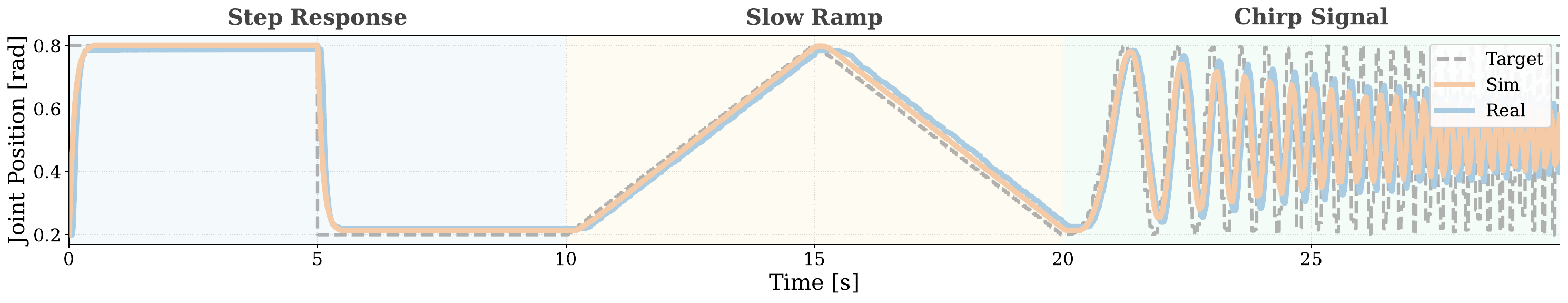}
    \vspace{-0.5cm}
    \caption{Actuator dynamics are aligned across simulation and the real world by executing a sequence of varying inputs to probe the system.}
    \label{fig:actuator_sysid}
\end{figure}

\section{System Identification of Actuator Dynamics}\label{appendix:actuator_sysid}
In IsaacSim, the key configurable actuators parameters include stiffness, damping, and joint friction.
To find the set of simulation parameters that results in actuation behaviors closely-aligned to the real hardware, we designed a series of trajectory sequences to probe the dynamic responses of each actuator, including step inputs, slow linearly-varying ramp inputs, and chirp inputs with increasing frequency, as shown in Fig.~\ref{fig:actuator_sysid}.
Then, we choose a fixed set of actuator parameters, apply onto the real robot, and execute the trajectory sequences and record the trajectory data consisting of joint angles at all time steps.
Next, we begin an automatic tuning process in simulation via a Bayesian optimization loop.
At each iteration, the current sampled joint parameters are applied to the simulated robot, the same trajectory sequences are executed, and the trajectory data is recorded.
We then compute the MSE loss between the sim and real trajectories and use it as the minimization objective for the Bayesian optimizer to sample the parameters in the next iteration.
Fig.~\ref{fig:actuator_sysid} shows an example of trajectory alignment after optimizing for the actuator parameters of a given joint.

\begin{figure}[t]
    \centering
    \begin{subfigure}{0.47\textwidth}
        \centering
        \includegraphics[width=\linewidth]{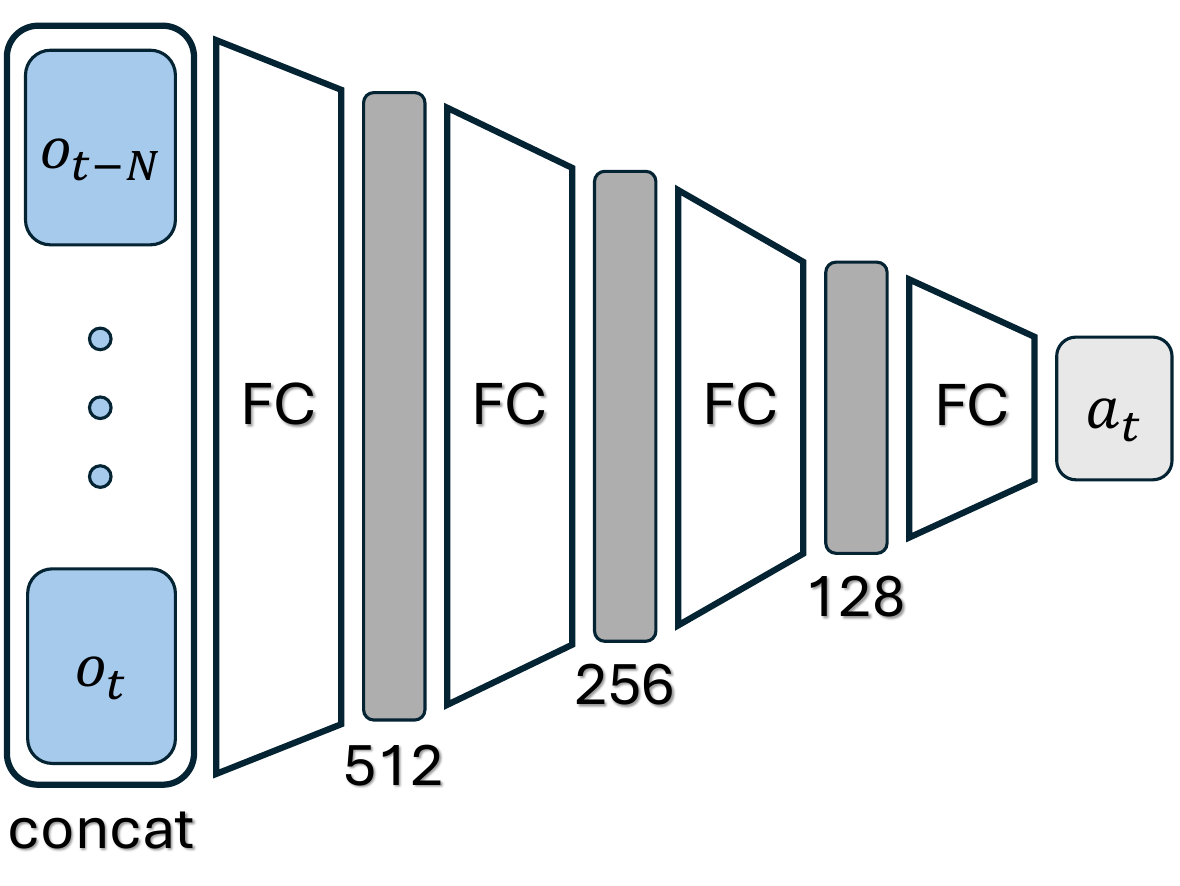}
        \caption{The past $N$-step observations are concatenated and passed as input to the MLP policy network.}
        \label{fig:mlp_arch_visualization}
    \end{subfigure}
    \hfill
    \begin{subfigure}{0.49\textwidth}
        \centering
        \includegraphics[width=\linewidth]{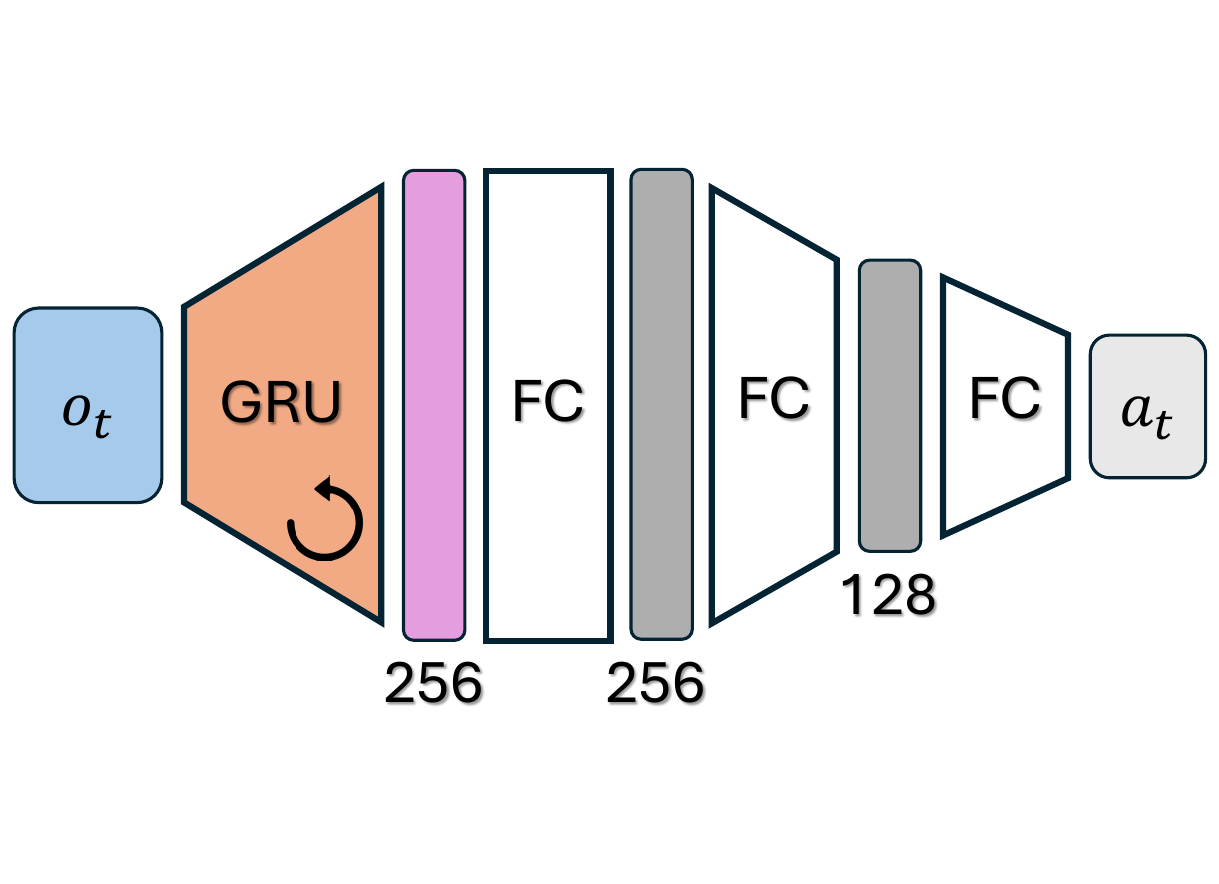}
        \caption{The observation from only the current step is passed as input to the recurrent policy network.}
        \label{fig:rec_arch_visualization}
    \end{subfigure}
    \caption{Comparison of policy network architectures, including (a) direct MLP architecture consisting of only fully-connected (FC) layers and (b) recurrent architecture consisting of a recurrent (GRU) layer followed by FC layers.}
    \label{fig:arch_visualization}
\end{figure}

\section{Policy Network Architecture Comparison}\label{appendix:policy_arch_comparison}
We analyze the performance of our recurrent policy architecture consisting of a single recurrent layer (GRU) which outputs a 256-dim latent that is passed to two linear layers with sizes $[256, 128]$ and ELU activation, and compare it against a direct MLP with three linear layers with sizes $[512, 256, 128]$ and ELU activation (see Fig.~\ref{fig:arch_visualization}).
Moreover, to examine whether providing longer temporal context via stacking more history observations to a MLP policy improves its performance, we train several policies conditioned on different numbers ($N=\{3, 5, 10, 20\}$) of history steps of concatenated observation vectors.
The PPO hyperparameters and training configurations are kept identical for the recurrent and MLP policies to ensure fairness (see Appendix~\ref{appendix:ppo_config}).
As shown in Fig.~\ref{fig:arch_comparison}, the performance gains saturate for MLP policies conditioned on longer observation histories.
For both tasks, we found that the recurrent policy results in better sample efficiency and convergence quality than the MLP policy.
One possible underlying factor for this result is that for incorporating longer-horizon temporal information, the recurrent layer maintains the same observation space dimension, whereas the direct MLP network requires concatenating larger numbers of past observation vectors and thus drastically increases the observation space dimension.
Another possible factor may be that the recurrent layer's update mechanism fuses the current observations with its internal hidden state, which may act as a ``regularizer" and effectively smooths out noisy observations.
\begin{figure}[ht]
    \centering
    \begin{subfigure}{0.49\textwidth}
        \centering
        \includegraphics[width=\linewidth]{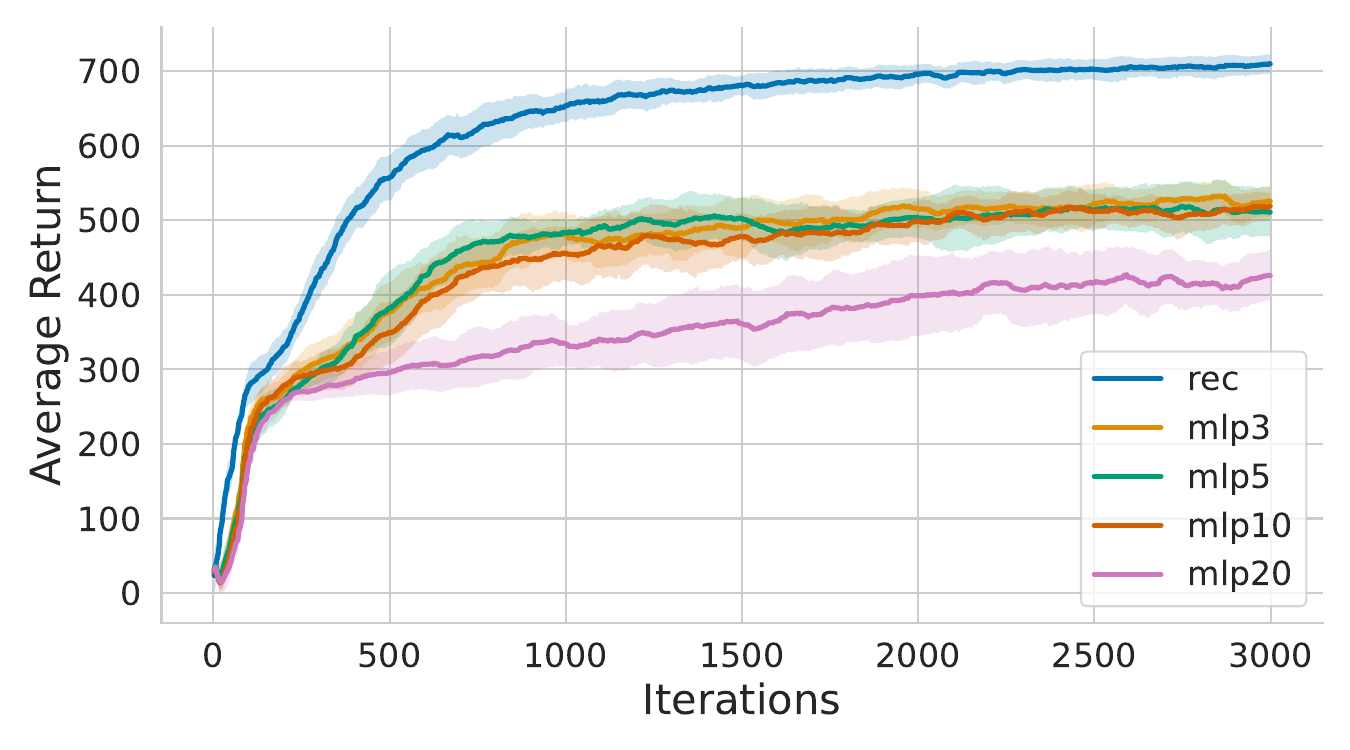}
        \caption{Peg-in-hole insertion.}
        \label{fig:insertion_arch_comparison}
    \end{subfigure}
    \hfill
    \begin{subfigure}{0.49\textwidth}
        \centering
        \includegraphics[width=\linewidth]{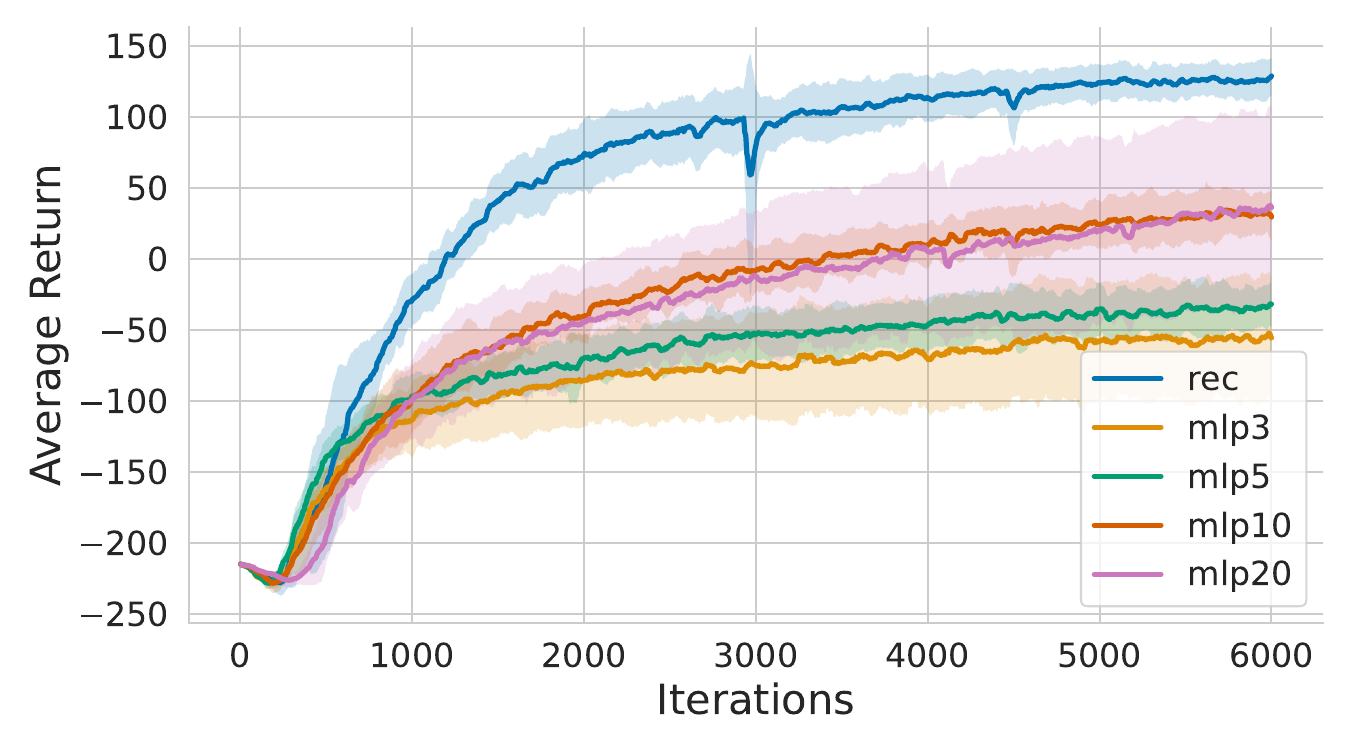}
        \caption{Ball balancing.}
        \label{fig:balance_arch_comparison}
    \end{subfigure}
    \caption{Comparing the training performance of our recurrent policy network against MLP networks with different history lengths of stacked input observation vectors as input. Shaded area represents the standard deviation over 5 random seeds.}
    \label{fig:arch_comparison}
\end{figure}

\section{RL Configuration \& Training Details}\label{appendix:rl_training_details}

\subsection{Actor \& Critic Observations}
We employ PPO with asymmetric actor and critic observations for policy training. 
At each time step, the policy observation includes the current joint angles $q_t$, previous actions $a_{t-1}$, and the contact representation.
For contact representations such as \textit{cop} and \textit{taxel} which are stored as higher-order tensors, we flatten them into a single vector and concatenate it to the proprioception vector.
The critic observation includes privileged task-relevant information in addition to the actor observations, and are summarized in Table~\ref{table:critic_observations}.
\vspace{-0.2cm}
\begin{table}[h]
\centering
\footnotesize
\caption{Summary of privileged critic observations for both tasks.}
\label{table:critic_observations}
\begin{tabular}{cc|cc}
\hline
\multicolumn{2}{c|}{Peg-in-Hole Insertion} & \multicolumn{2}{c}{Ball Balancing} \\
observation     & quantity    & observation    & quantity   \\ \hline
    peg position     &  $p_{\rm{peg}}\in\mathbb{R}^3$    & plate position               &   $p_{\rm{plate}}\in\mathbb{R}^3$         \\
    peg rotation    & $R_{\rm{peg}}\in\mathbb{R}^4$    &   plate rotation             &   $R_{\rm{plate}}\in\mathbb{R}^4$        \\
    goal vector  & $r_{\rm{goal}}=p_{\rm{goal}}-p_{\rm{peg}}\in\mathbb{R}^3$  &  plate linear velocity  &  $v_{\rm{plate}}\in\mathbb{R}^3$ \\
    goal reached  & $\mathds{1}(\|r_{\rm{goal}}\|\leq \epsilon)$   & plate angular velocity & $\omega_{\rm{plate}}\in\mathbb{R}^3$ \\
      &  & ball position   & $p_{\rm{ball}}\in\mathbb{R}^3$  \\
       &   & ball linear velocity    &  $v_{\rm{ball}}\in\mathbb{R}^3$ \\
         &   & goal distance  &  $d_{\rm{goal}}=p_{\rm{ball}}-p_{\rm{plate}}\in\mathbb{R}^3$ \\ \hline
    
\end{tabular}
\end{table}

\subsection{Policy Action Space}
At each time step, the policy network outputs 16-dim actions $a\in\mathbb{R}^{16}$, which correspond to target joint position \textit{increments} across all 16 DOFs.
We first clip the actions to the range $[-1, 1]^{16}$, and apply an action scale of 0.03 for peg-in-hole insertion and 0.05 for ball balancing.
Then, we apply an exponential moving average (EMA) with $\alpha=0.5$ to the scaled target joint position increments, and add it to the previously commanded joint target positions to obtain the current joint targets which are then tracked by the PD controller.
For the PD controller gains, we use $P=3.0, D=0.1$ for the insertion task for more compliant control, and $P=6.0, D=0.15$ for the balancing task for more reactive control.
\subsection{Rewards}
The reward terms for each task consists of task progress-based reward terms and regularizing penalty terms, and are summarized in Table~\ref{table:rewards}.
\begin{table}[h]
\renewcommand{\arraystretch}{1.2}
\centering
\footnotesize
\caption{Summary of reward terms for both tasks.}
\label{table:rewards}
\begin{tabular}{ccc}
\hline
Reward Term & Expression & Weight \\ \hline
\multicolumn{3}{c}{Peg-in-Hole Insertion} \\ \hline
goal distance reward & $\rm{exp}(-0.5(d_{\rm{goal}}/0.015))$ & 1.0    \\
goal reached reward (given once) & $\mathds{1}(\|r_{\rm{goal}}\|\leq\epsilon)$ & 400.0 \\
good contact reward & $\sum_{i=1}^{N_{\rm{sensors}}}\mathds{1}(\|f_i\|\geq1.0)$ & 0.25 \\
rotation deviation (from z-axis) penalty & $1 - \rm{exp}(-0.5(\|R_{\rm{peg,z}}\|/0.6)^2)$ & 1.0 \\
hand DOF pos deviation penalty & $1 - \rm{exp}(-0.5(\|q-q_{\rm{default}}\|/0.2)^2)$ & 1.0 \\ \hline
\multicolumn{3}{c}{Ball Balancing} \\ \hline
goal distance reward & $\rm{exp}(-d_{\rm{goal}}/0.05)$ & 1.0    \\
plate contact reward & $\sum_{i=1}^{N_{\rm{sensors}}}\mathds{1}(\|f_i\|\geq0)$ & 0.2 \\
ball-plate relative linear velocity penalty & $\rm{tanh}(\|v_{\rm{ball}}-v_{\rm{plate}}\|/0.3)$ & 1.0 \\
plate position deviation penalty & $\rm{tanh}(\|p_{\rm{plate}}-p_{\rm{plate,default}}\|/0.15)$ & 1.0 \\
plate yaw penalty & $\rm{tanh}(\|R_{\rm{plate,z}}\|/0.5)$ & 1.0 \\
ball fallen penalty & $\mathds{1}(p_{\rm{ball,z}}\leq0.2)$ & 200.0 \\
action diff penalty & $\rm{tanh}(\|a_{t}-a_{t-1}\|^2/16.0)$ & 1.0 \\ \hline

\end{tabular}
\end{table}

\subsection{Domain Randomization}
We employ systematic domain randomization to object physical properties including friction and mass, object states during resets, and observation noise and delay to make the policy robust against noisy real-world variations.
Domain randomization setups for both tasks are summarized in Table~\ref{table:domain_randomization}.
\setlength{\tabcolsep}{20pt}
\begin{table}[h]
\renewcommand{\arraystretch}{1.2}
\centering
\footnotesize
\caption{Domain randomization setup including both task-specific and general configurations.}
\label{table:domain_randomization}
\begin{tabular}{lc}

\hline
\multicolumn{2}{c}{Peg-in-Hole Insertion} \\ \hline
Peg: Mass (kg)  & $[0.03, 0.04]$ \\
Peg: Friction (static \& dynamic) & $[0.2, 0.4]; [0.1, 0.2]$ \\
Peg: Initial Roll \& Pitch (rad) & $+\mathcal{U}(-0.2, 0.2)$ \\
Peg: Initial Yaw (rad) & $+\mathcal{U}(0, 2\pi)$ \\
Hole: Friction (static \& dynamic) & $[0.3, 0.5]; [0.1, 0.3]$ \\
Hand: Friction (static \& dynamic) & $[0.5, 0.7]; [0.3, 0.5]$ \\ 
Hand: Initial Position (cm) & $+\mathcal{U}(-0.5, 0.5)$   \\
Hand: Initial Roll \& Pitch (rad) & $+\mathcal{U}(-0.02, 0.02)$ \\ 
Hand: Initial DOF pos (rad) & $+\mathcal{U}(-0.05, 0.05)$ \\ \hline
\multicolumn{2}{c}{Ball Balancing} \\ \hline
Ball: Mass (kg)  & $[0.05, 0.25]$ \\
Ball: Friction (static \& dynamic) & $[0.01, 0.02]; [0.0, 0.01]$ \\
Ball: Initial Position (cm) & $+\mathcal{U}(-5.0, 5.0)$ \\
Plate: Mass (kg) & $[0.035, 0.055]$ \\
Plate: Friction (static \& dynamic) & $[0.01, 0.02]; [0.0, 0.01]$ \\
Plate: Initial Position (cm) & $+\mathcal{U}(-0.5, 0.5)$ \\
Hand: Friction (static \& dynamic) & $[1.9, 2.0]; [1.8, 1.9]$ \\
Hand: Initial Roll \& Pitch (rad) & $+\mathcal{U}(-0.02, 0.02)$ \\ 
Hand: Initial DOF pos (rad) & $+\mathcal{U}(-0.05, 0.05)$ \\ \hline
\multicolumn{2}{c}{General} \\ \hline
PD Controller: P Gain & $\times\mathcal{U}(0.8, 1.2)$ \\
PD Controller: D Gain & $\times\mathcal{U}(0.7, 1.3)$ \\ \hline
Joint Pos Obs Noise (rad) & $+\mathcal{U}(-0.1, 0.1)$ \\
Force Vector Obs Noise: Prob & $0.2$ \\
Force Vector Obs Noise: Rotation (rad) & $+\mathcal{U}(-0.1, 0.1)$ \\
Force Vector Obs Noise: Magnitude & $\times\mathcal{U}(0.9, 1.1)$ \\
Contact Pos Obs Noise Prob & $0.2$ \\
Contact Pos Obs Noise (cm) & $+\mathcal{U}(-0.1, 0.1)$ \\ \hline
Joint Pos Obs Delay (s) & $0.05$ \\
Contact Obs Delay (s) & $[0.05, 0.1]$ \\ \hline

\end{tabular}
\end{table}
\subsection{PPO Configuration \& Training Results}\label{appendix:ppo_config}
\begin{figure}[ht]
    \centering
    \begin{subfigure}{0.49\textwidth}
        \centering
        \includegraphics[width=\linewidth]{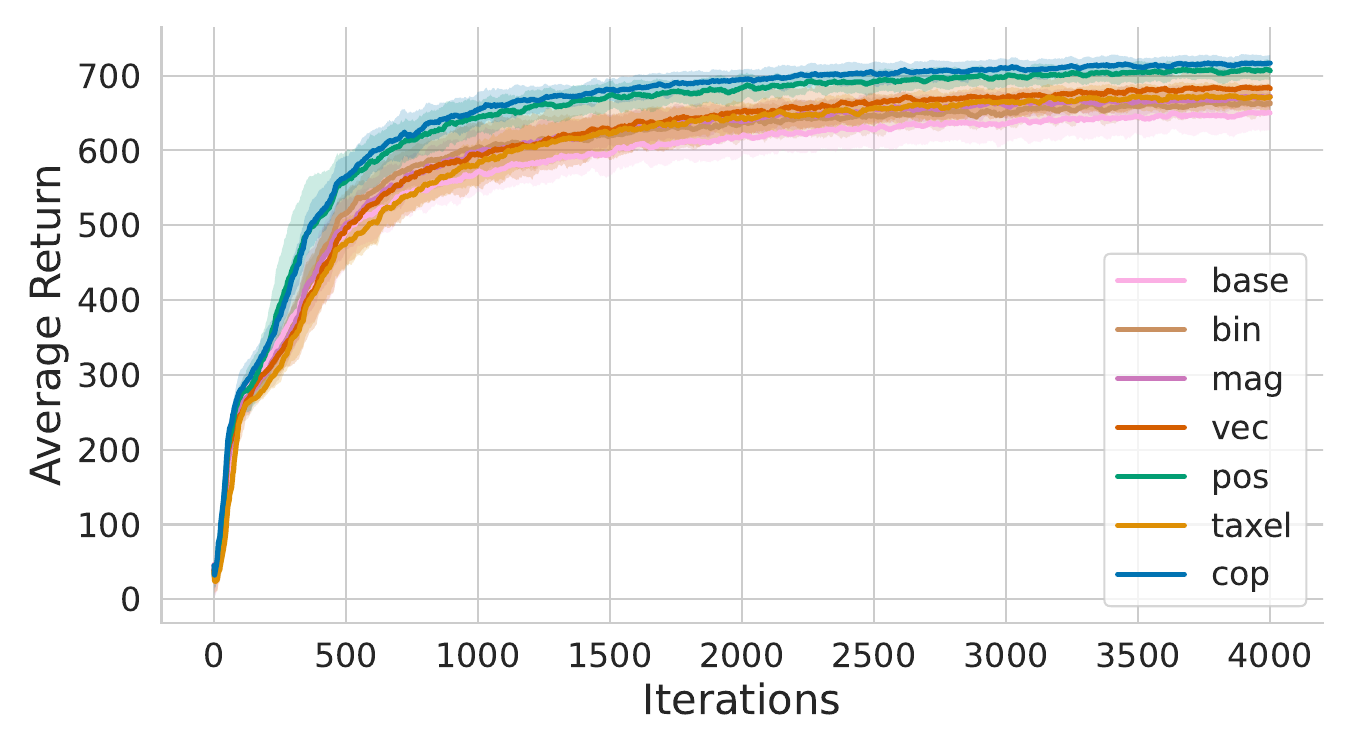}
        \caption{Peg-in-hole insertion.}
        \label{fig:insertion_obs_comparison}
    \end{subfigure}
    \hfill
    \begin{subfigure}{0.49\textwidth}
        \centering
        \includegraphics[width=\linewidth]{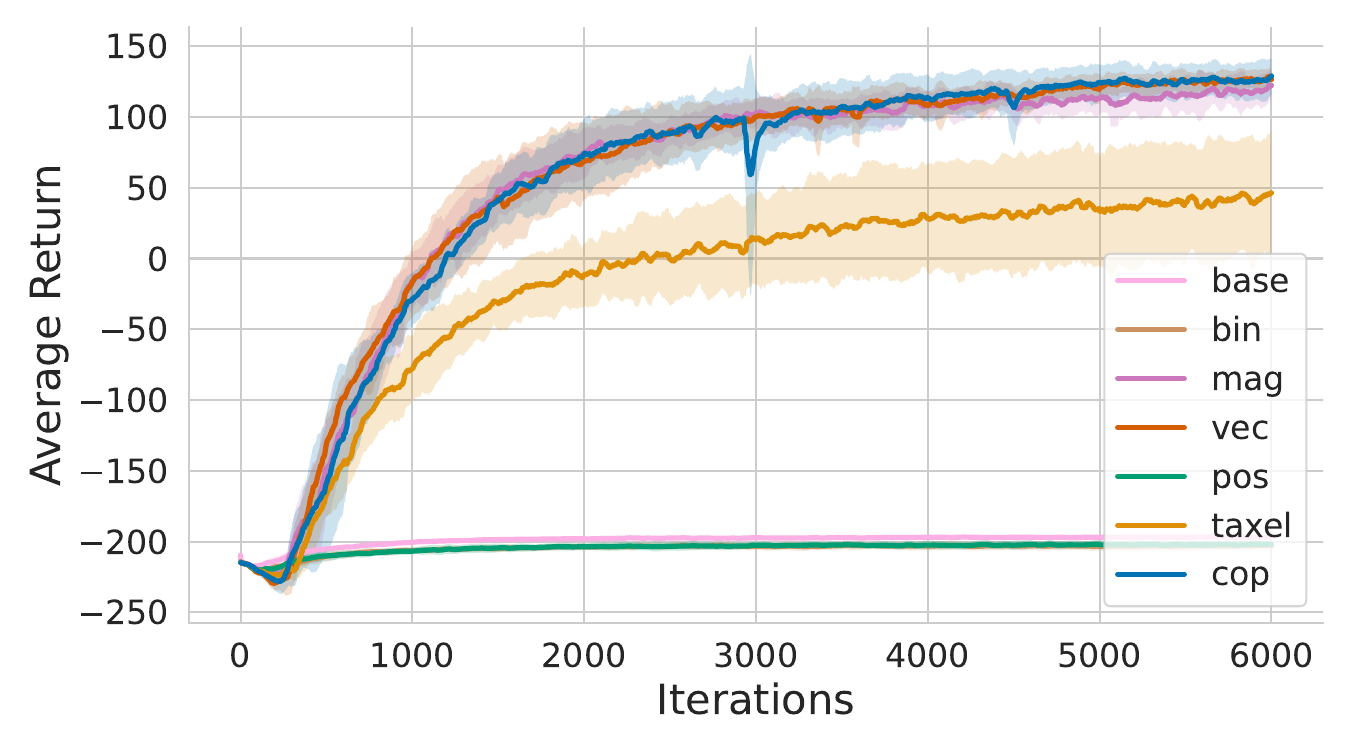}
        \caption{Ball balancing.}
        \label{fig:balance_obs_comparison}
    \end{subfigure}
    \caption{Comparison of training performance across various contact representations for policy observation. Shaded area represents the standard deviation over 5 random seeds.}
    \label{fig:obs_comparison}
\end{figure}

We use the Proximal Policy Optimization (PPO) algorithm to learn RL policies.
To ensure stable learning across varying feature scales, we apply online observation normalization to both the actor and critic inputs. 
The training uses an adaptive learning rate schedule starting at $5.0 \times 10^{-4}$ with a target KL divergence of $0.016$. 
During each iteration, we collect 64 steps for the insertion task and 16 steps for the balancing task per environment and perform 5 epochs of updates over 4 mini-batches.
We use discount factor $\gamma = 0.99$, GAE parameter $\lambda = 0.95$, clipping range of $0.2$, and entropy coefficient of $0.005$ to encourage exploration.
Fig.~\ref{fig:obs_comparison} shows the training results for both tasks.
In the peg-in-hole insertion task, the \textit{cop} policy on average attains the highest rewards during training than all other observations, though the difference is fairly marginal.
For this task, the high exploration noise during training may have led to successful insertions through action stochasticity, thus artificially inflating the rewards of less robust policies. 
In contrast, as shown in the real-world transfer results, during inference, the cop observation enables the policy to achieve more consistent and robust performance than other observations.
For the ball balancing task, the policies conditioned on contact representation which lack explicit force information, including \textit{base}, \textit{bin}, and \textit{pos}, completely fails to learn the task.
Consistent with the real-world results, we observe that the \textit{taxel} observation results in worse performance than more compact contact representations such as \textit{force} and \textit{cop}, which is likely due to the high dimensionality and complexity of the raw taxel forces.

\section{Insertion Task Asset Details}\label{appendix:insertion_asset_details}
\begin{figure}[ht]
    \centering
    \includegraphics[width=1.0\textwidth]{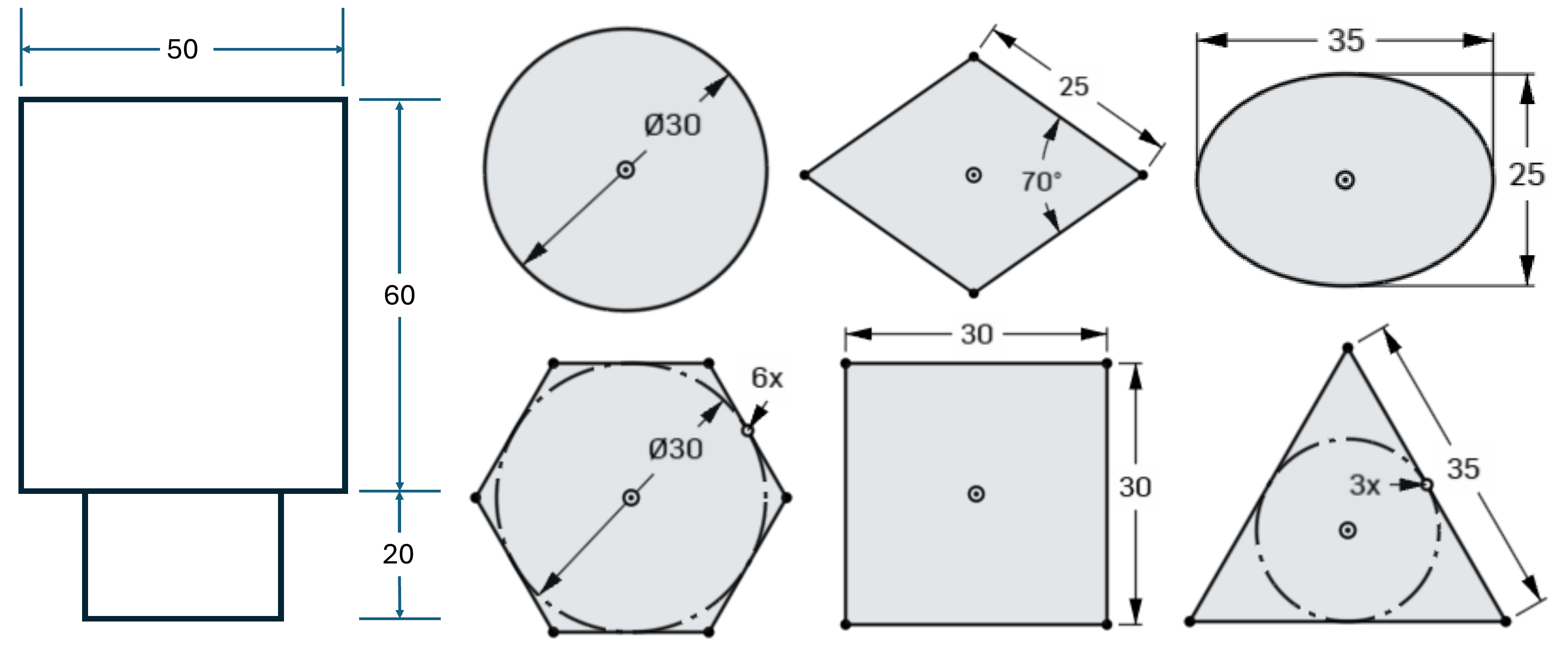}
    \caption{\textbf{Left}: The dimensions of the insertion peg which are shared across all six shapes. \textbf{Right}: The dimensions of each insertion shape. All lengths are given in millimeters (\si{\mm}).}
    \label{fig:insertion_peg_dimensions}
    \vspace{-0.5cm}
\end{figure}

Fig.~\ref{fig:insertion_peg_dimensions} details the dimensions of the insertion peg assets across the six different shapes -- circle, diamond, ellipse, hexagon, square, and triangle.
For the corresponding hole assets, the sizes of the hole are increased by 10\% in the $x$ and $y$ axes, resulting in 10\% error tolerance for insertion.
This was motivated by the observation that in simulation, using zero or lower tolerance induces ``jamming" behaviors between the head of the insertion peg and the hole during the insertion process, likely due to mesh approximations and potential numerical errors in the physics engine's collision resolution.

\end{document}